%% file: main.tex
\documentclass[]{gtech}
\PassOptionsToPackage{table, usenames, dvipsnames}{xcolor}
\usepackage{xcolor}
\usepackage{amssymb}
\usepackage{multirow}
\usepackage{bigdelim}
\usepackage{todonotes}
\usepackage{longtable}
\usepackage{tabularray}
\usepackage{wrapfig}
\usepackage[most]{tcolorbox} 
\usepackage{url}
\usepackage{float}
\usepackage{datatool}
\usepackage{enumitem}
\usepackage{subcaption} 
\usepackage[justification=centering]{caption}

\usepackage{changes} 
\usepackage{amsmath}

\RequirePackage{tgpagella} 
\RequirePackage{mathpazo}  
\RequirePackage{inconsolata} 
\usepackage{makecell}

\renewcommand{\title}[1]{\newcommand{\titlelist}{{\huge\fontfamily{optimistic}\selectfont #1}}}

\newcommand{\model}{\texttt{Ring-lite}}
\newcommand{\distillmodel}{\texttt{Ring-lite-distill}}
\newcommand{\qwenbasedistill}{\texttt{Ring-distill-Qwen-7B}}
\newcommand{\dsbasedistill}{\texttt{DeepSeek-R1-Distill-Qwen-7B}}
\definecolor{prompt}{HTML}{5f84e4}
\definecolor{img}{HTML}{820100}

\usepackage{arydshln}
\definecolor{CQColor}{rgb}{0.0,0.0,1.0} 

\usepackage{colortbl}
\usepackage{amssymb}
\usepackage{pifont}
\usepackage{booktabs,multirow}
\usepackage{makecell}
\usepackage{tabulary}
\usepackage{fontawesome5}
\usepackage{bbding}
\usepackage{multicol}

\newlength\savewidth

\title{\textcolor[HTML]{0369ff}{Ring}-lite: Scalable Reasoning via C3PO-Stabilized Reinforcement Learning for LLMs}

\author[*]{Ling Team, Inclusion AI}

\contribution[*]{See Contributions section (Sec.~\ref{sec:contri}) for full author list.}

\abstract{\fontsize{11pt}{12pt} \textit{We present \model{}, a Mixture-of-Experts (MoE)-based large language model optimized via reinforcement learning (RL) to achieve efficient and robust reasoning capabilities. 
Built upon the publicly available Ling-lite model, a 16.8 billion parameter model with 2.75 billion activated parameters, our approach matches the performance of state-of-the-art (SOTA) small-scale reasoning models on challenging benchmarks (e.g., AIME, LiveCodeBench, GPQA-Diamond) while activating only one-third of the parameters required by comparable models. 
To accomplish this, we introduce a joint training pipeline integrating distillation with RL, revealing undocumented challenges in MoE RL training.
First, we identify optimization instability during RL training, and we propose \underline{C}onstrained \underline{C}ontextual \underline{C}omputation \underline{P}olicy \underline{O}ptimization(\textbf{C3PO}), a novel approach that enhances training stability and improves computational throughput via algorithm-system co-design methodology. 
Second, we empirically demonstrate that selecting distillation checkpoints based on \textbf{entropy loss} for RL training, rather than validation metrics, yields superior performance-efficiency trade-offs in subsequent RL training. 
Finally, we develop a \textbf{two-stage} training paradigm to harmonize multi-domain data integration, addressing domain conflicts that arise in training with mixed dataset. We will release the model, dataset, and code.}}
\date{Jun 17, 2025\vspace{-1mm}}
\gtechdata[Code]{\url{https://github.com/inclusionAI/Ring}}

\begin{document}
\maketitle

\input{sections/intro}

\input{sections/data}

\input{sections/method}
\input{sections/exp}

\input{sections/conclu}
\input{sections/author}

\bibliographystyle{assets/plainnat}
\bibliography{main}

\end{document}

%% file: sections/intro.tex
\begin{figure*}[!hbt]
\centering
\includegraphics[width=0.98\textwidth]{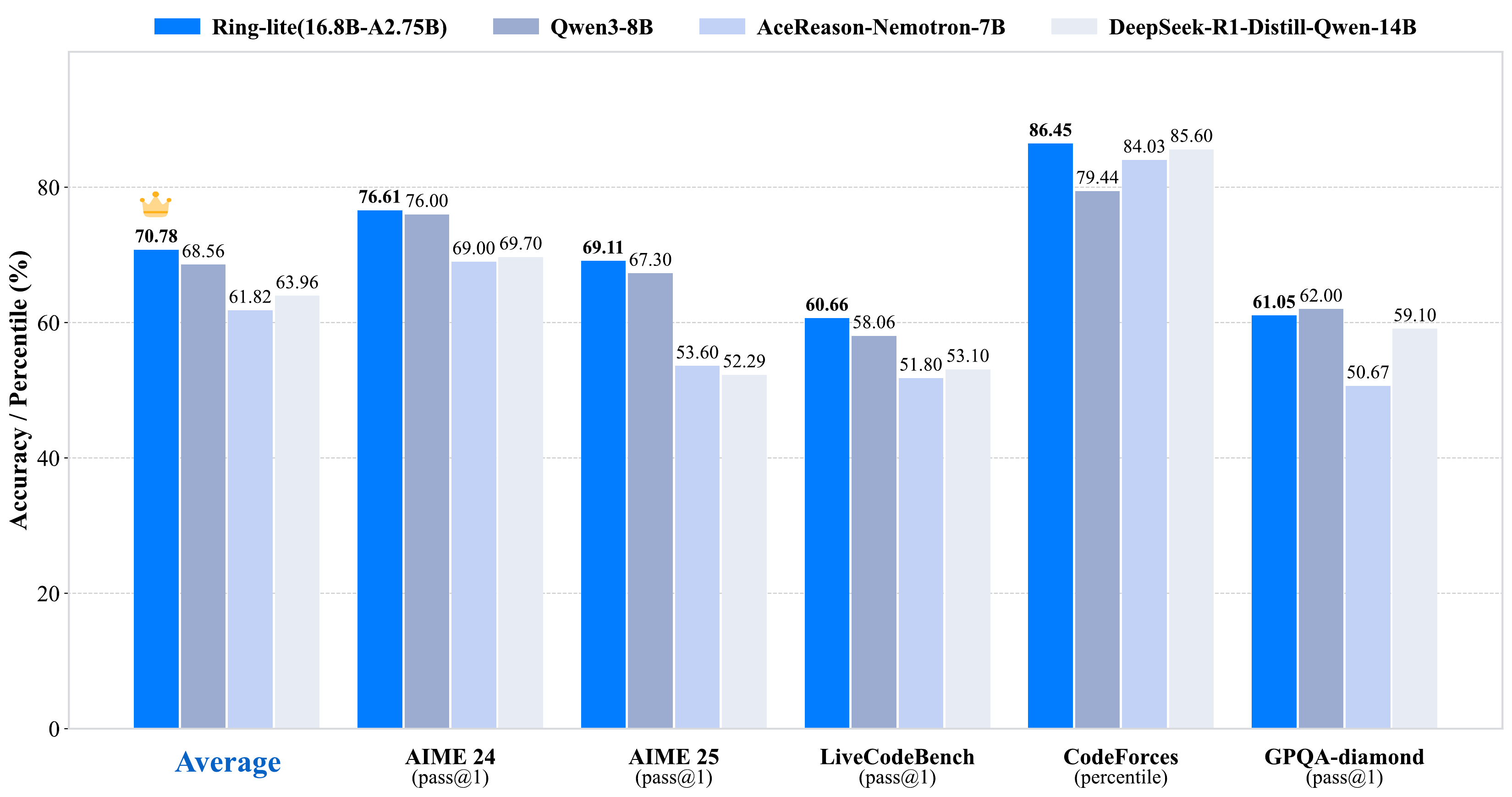}
\caption{\textbf{Benchmark performance of Ring-lite}}
\label{fig:ring-lite-performance}
\end{figure*}

\section{Introduction}
\label{sec:intro}

The remarkable success of the OpenAI-O1 series models~\citep{openai2024openaio1card} and DeepSeek-R1~\citep{deepseekai2025deepseekr1} has demonstrated the substantial potential of large-scale reinforcement learning (RL) for complex reasoning tasks, attracting significant research attention. However, the detailed methodologies employed in the training of these models have not been fully disclosed, creating a significant knowledge gap that hinders further advancements in the field. This lack of transparency in training techniques and strategies impedes the reproduction and extension of these results by other researchers. 

Recent advancements, such as DAPO~\citep{yu2025dapo}, Open-Reasoner-Zero~\citep{hu2025orz}, and DeepCoder~\citep{deepcoder2025}, have demonstrated competitive reasoning performance through task-specific RL strategies, accompanied by publicly released models and datasets. However, their contributions remain narrowly scoped, focusing predominantly on isolated domains such as mathematics or code generation, with limited cross-task generalization.
Furthermore, while current research has largely focused on dense model architectures~\citep{yu2025dapo,zhang2025srpocrossdomainimplementationlargescale}, scant attention has been devoted to exploring the potential of Mixture of Experts (MoE)~\citep{deepseekai2025deepseekr1,ling} paradigms in this context. 
Of particular concern is the persistent challenge of training stability—a fundamental prerequisite for scaling RL-based reasoning systems that remains systematically unaddressed. The dynamic interaction between specialized experts in MoE architectures introduces complex gradient synchronization and parameter update conflicts, often manifesting as oscillating loss or disaster forgetting of acquired reasoning skills. Without robust methodologies to ensure stable training in large-scale RL training, the theoretical advantages of MoE frameworks cannot be fully realized in practice.

In this work, we introduce \model{}, a fully open-source Mixture of Experts (MoE) reasoning model designed to enhance multi-domain reasoning capabilities built upon the publicly available Ling-lite model~\citep{ling}. 
To the best of our knowledge, this is the first work that integrates an open training framework, open training data, and an open model, specifically targeting the domains of mathematics, coding, and STEM. Furthermore, \model{} systematically delves into the instability issues prevalent in RL training and the conflicts arising from capability integration across domains. To solve the instability problem, we propose \underline{C}onstrained \underline{C}ontextual \underline{C}omputation \underline{P}olicy \underline{O}ptimization(\textbf{C3PO}), a novel token-level optimization framework for reinforcement training. 
The experimental results obtained by our model on complex reasoning tasks, coupled with its more stable reward curves compared to the widely-used conventional Group Relative Policy Optimization (GRPO) method, substantiate the efficacy of our approach.
Our model, \model{} with 16.8B total parameters and 2.75B activation parameters, establishes state-of-the-art performance across mathematical reasoning, code generation, and STEM problem-solving benchmarks, demonstrating superior performance by matching or surpassing dense models with under 10B parameters, the standard baseline for comparable architectures according to the scaling law analyses~\citep{ling}. To our knowledge, \model{} represents the first publicly available mixture-of-experts (MoE) system operating at this parameter-efficiency frontier, serving both as a methodological innovation in efficient architectural design through dynamic sparse activation and as an open-access resource that lowers barriers to cutting-edge AI research. This dual contribution advances both theoretical understanding of neural scaling laws and practical democratization of high-performance language model technologies.
Specifically, \model{} achieves impressive scores of 76.61\% and 69.11\% on AIME2024 and AIME2025, two challenging math competition-style benchmarks, 60.66\% and 86.45\% on LiveCodeBench and Codeforces, two challenging code contest benchmarks for code generation, 61.05\% on GPQA-diamond, the graduate-Level science QA benchmark. It surpasses Qwen3-8B~\citep{qwen3} in average and approaches the performance of other top-tier reasoning models.

In short, our work marks a pivotal advance in the democratization of AI research and development, as it provides the broader community with a fully open-source solution. By doing so, our work significantly lowers the barrier to entry for exploring multi-domain reasoning, empowering researchers and practitioners alike to contribute to and benefit from this burgeoning field. 
The main contributions are summarized as follows.
\begin{itemize}
    \item For the first time, we open-source a multi-domain MoE reasoning model~\footnote{https://github.com/inclusionAI/Ring}, encompassing a open source infrastructure (framework), training methodologies, and training datasets. The entire transparent training pipeline is detailed, including Long-CoT Supervised Fine-Tuning (SFT) and reasoning-specific reinforcement learning (RL).
    \item We identify a critical challenge in the training instability of reasoning models and propose C3PO, a framework that implements a fixed training token size (budget) to eliminate response length variance and select high-entropy base models to stabilize learning dynamics. The framework integrates a reinforcement learning methodology grounded in an algorithm-engineered co-design paradigm, thereby not only ensuring long-term training stability but also achieving significant gains in computational efficiency.
    \item Spanning multiple domains (math, code and science), we observed inter-domain data conflict and introduced a capability integration method (stage-wise training and balanced data mixing) to address this issue.
\end{itemize}

The structure of this work proceeds as follows: Section~\ref{sec:data} details the dataset curation process, including data cleaning and filtering. Section~\ref{sec:method} systematically outlines the methodological framework, emphasizing its contributions and implementation specifics. Finally, Section~\ref{sec:bench} evaluates the model’s performance against established benchmarks and synthesizes critical insights gleaned from both quantitative results and qualitative observations.

%% file: sections/data.tex
\section{Data}
\label{sec:data}
Our training dataset comprises two components: (1) long Chain-of-Thought (Long-CoT) supervised fine-tuning (SFT) data, employed to train a cold-start model, and (2) reinforcement learning (RL) data, designed to enhance reasoning capabilities.

\subsection{Long-CoT Data}
\label{sec:long-cot-data}
To activate a base model's reasoning capability, a comprehensive dataset of high-quality samples exhibiting Long-CoT reasoning patterns was curated. The query pool was sourced from open-source repositories and further enriched through synthetic generation using large language models (LLMs). To ensure the production of high-fidelity responses with Long-CoT, we implemented an iterative refinement pipeline that synergistically combines automated model generation, expert manual annotation, and rejection sampling mechanisms. After that, rigorous data-cleansing protocols were applied, including detection and removal of repetitive patterns, mixed-language artifacts, and other noise sources, to yield a robust and high-quality dataset.
The final data is predominantly dominated by three major domains: Mathematics (64.5\%), Code (25.5\%), and Science (9.2\%, encompassing some high-quality and difficult samples generated by SHARP~\citep{wu2025SHARP}). The remaining portion of the dataset includes contributions from other categories, such as medicine and history domains. 
In short, our long-CoT SFT dataset enabled effective multi-domain reasoning (spanning mathematics, coding, and science), providing a robust initialization for subsequent reinforcement learning training. 

\subsection{RL Training Data}
\subsubsection{Domain-Specific Reasoning Data}
\begin{itemize}
    \item \textbf{Math} We begin by sourcing a wide range of mathematical problems to enrich the diversity and coverage of our math reasoning dataset. These problems are mainly obtained from two channels: open-source datasets and self-collected data. For open-source datasets, we included datasets that have been meticulously curated for reinforcement learning, including BigMath~\citep{albalak2025bigmath}, DeepScaleR~\citep{deepscaler2025}, DAPO~\citep{yu2025dapo}, DeepMath-103K~\citep{he2025deepmath103k}, etc. To further expand our dataset, we crawled online math forums and collected authentic school examinations. Specifically, we extracted problems from the contest section of the Art of Problem Solving (AoPS) website\footnote{https://artofproblemsolving.com/community/c13\_contest\_collections}, which archives comprehensive records of historical mathematics competitions from diverse regions and educational levels. Additionally, we gathered a wide range of human-written problems utilized in school exams and mathematics competitions across various educational stages, such as high school and college. This extensive process yielded an initial collection of more than tens of thousands of math problems. We then applied stringent data cleansing and filtering protocols to ensure quality and relevance, ultimately refining the dataset to include over 73,000 high-quality math problems suitable for our reinforcement learning processes.
    \item \textbf{Code} The dataset was curated from open-source programming competition resources, primarily drawn from CodeContest~\citep{codecontests}, TACO~\citep{taco} and APPS~\citep{hendrycksapps2021}, additionally some problems from the QOJ online judge platform~\footnote{{https://qoj.ac}}. To ensure data quality and training suitability, a multi-stage filtration process was implemented. First, test cases exhibiting format inconsistencies—such as erroneous line breaks or extraneous spaces—were systematically removed, along with truncated content marked by ellipses or incomplete patterns. Subsequently, all ``Accepted''(AC) solutions underwent rigorous validation through our code sandbox environment. This verification step eliminated submissions with unresolved external dependencies and discarded implementations that failed extended test cases due to computational inefficiencies (\textit{e.g.,} O(n$^2$) algorithms for n > 10$^5$) or memory overflow vulnerabilities. Output standardization was enforced by normalizing whitespace conventions and aligning floating-point precision thresholds across platforms to mitigate inconsistencies in evaluation criteria.
    To ensure integrity, only problems accompanied by at least one fully validated AC solution capable of passing all associated test cases were retained, thereby preserving practical problems with existing solution. Semantic deduplication was applied to remove overlapping problems from public coding benchmarks, minimizing the risk of evaluation bias through contamination control. The final curated dataset comprises approximately 14,000 code samples, each accompanied by verified executable solutions and rigorously validated test cases, establishing a robust foundation for reward computation and model training.
    \item \textbf{Science} For the science domain, our RL training data construction followed a three-stage evolution to ensure quality and difficulty alignment. Initially, we sourced open datasets such as Nemotron-CrossThink ~\citep{akter2025nemotron0crossthink0} and SCP-116K ~\citep{lu2025scp116khighqualityproblemsolutiondataset}, etc., to establish a baseline for scientific reasoning. As model capabilities improved, we employed the SHARP ~\citep{wu2025SHARP} synthesis pipeline to generate harder, verifiable problems. However, due to the difficulty ceiling and verification limitations of synthetic and open-source data, our final RL training relied exclusively on a third-stage dataset. This consisted of a proprietary collection of high-difficulty, human-annotated science problems drawn from advanced natural science domains. Sources included Olympiad competitions and graduate-level (Master’s and PhD) exams. We then applied a rigorous curation process—encompassing quality filtering, answer verification, and domain-specific tagging—resulting in a refined set of 3,833 high-quality scientific problems suitable for reinforcement learning.
\end{itemize}

\subsection{Data Curation}
The efficacy of the reinforcement learning process is heavily dependent on the quality of the training datasets. Through our initial investigations, we discover that data contamination issue persists even in the widely adopted open-source datasets. To ensure a high-quality training dataset for reinforcement learning, we developed an extensive and rigorous data curation pipeline, comprising several stages designed to ensure the complete decontamination of our data, thus making it readily prepared for RL training. The core components of our data processing protocol are primarily divided into the following three critical phases:

\begin{figure*}[ht]
\centering
\includegraphics[width=1.0\textwidth]{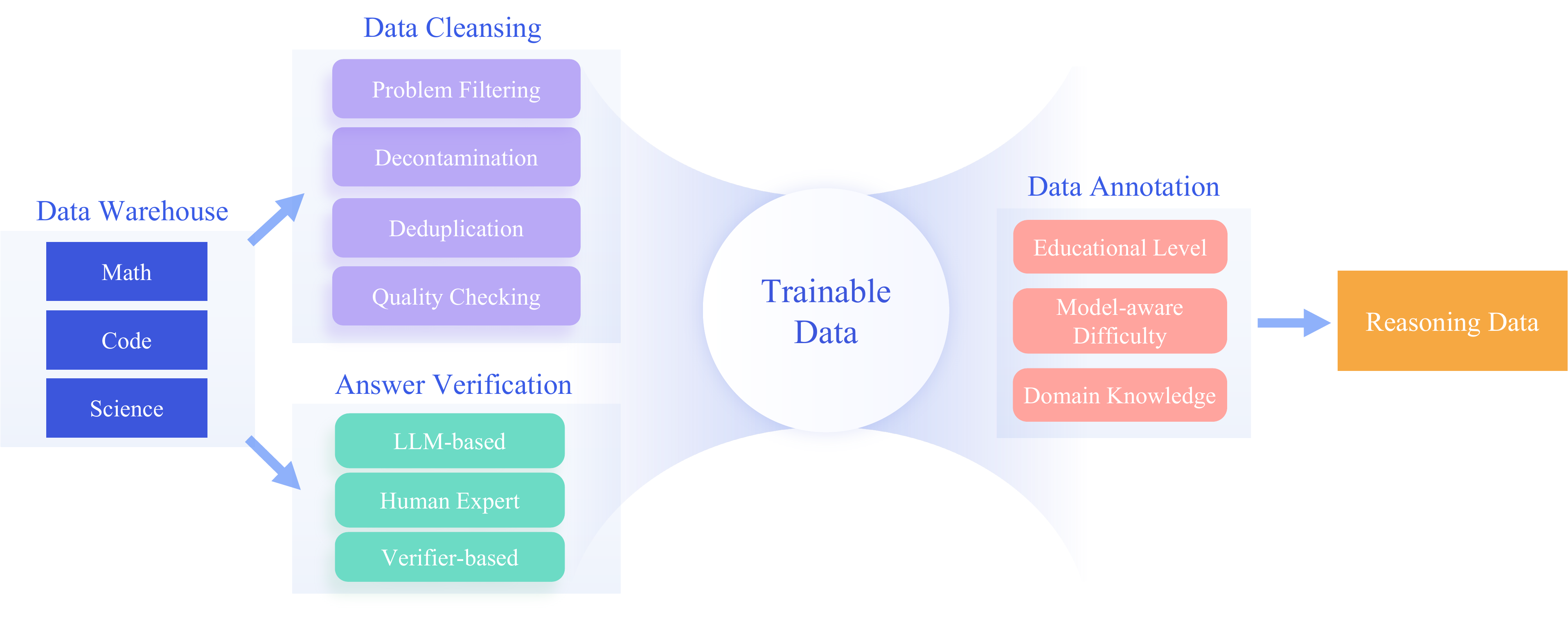}
\caption{\textbf{The data curation pipeline of \model{}}}
\label{fig:data-pipeline}
\end{figure*}
\paragraph{Data Cleansing} We first exclude problems with invalid character, images, multi-subquestions, and those lacking valid answers. We conducted strict both character-based and semantic-based deduplication and decontamination on the dataset to ensure strict data cleansing. We also remove problems which cannot be uniquely sovled or susceptible to be easily guessed, such as multiple-choice questions, and problems that can be answered with True/False, Yes/No, etc.
\paragraph{Answer Verification} To ensure the correctness of answers associated with problems in our dataset, we conduct thorough verification using diverse approaches. Specifically, we employ an LLM-based method to assess the quality of each answer. We utilize LLMs of different sizes to generate multiple individual solutions for each problem. Based on the verifiers used in RL training, we calculate the model-aware pass rate. Additionally, we engage human experts to manually annotate the answers. Problems that do not pass either verification method are excluded from our dataset.
\paragraph{Data Annotation} To optimize the data selection strategy, we meticulously annotate each reasoning problem. Specifically, each problem is labeled with multi-dimensional attributes, such as data source, educational level, domain-specific knowledge, and more. For instance, we use the Mathematical Subject Classification (MSC) categories to assess the themes of our math problems. Additionally, we provide model-aware difficulty by computing the solve rate based on our distilled model. Problems that receive all correct solutions are deemed inefficient for RL training; therefore, we remove those problems. Conversely, problems that are unsolvable by both our distilled model and DeepSeek-R1 are also discarded to ensure that the remaining data contribute effectively to policy gradient updates in reinforcement learning.

%% file: sections/method.tex
\section{Method}
\label{sec:method}
\subsection{Preliminary}
\label{sec:preliminary}
\textbf{G}roup \textbf{R}elative \textbf{P}olicy \textbf{O}ptimization (\textbf{GRPO}) algorithm is widely used such as DeepSeek-R1, Qwen3 and so on. For each question-answer pair $(q,a)$ in the training dataset $\mathcal{D}$, we generate 
$K$ responses (i.i.d.) through the policy model $\pi_{\theta_\text{old}}$. The reward $R_i$ of the response $y_i$ is determined by the reward model or rule-based verifier.
GRPO estimates the advantage via group-normalized rewards instead of the value model, ${A}_{i,t} = \frac{R_i - \text{mean}(\{R_i\}_{i=1}^K)}{\text{std}(\{R_i\}_{i=1}^K)}$. Specifically, the GRPO loss is formulated as:
\begin{equation}
\begin{aligned}
\mathcal{L}_\text{GRPO}(\theta)& = -\mathbb{E}_{(q,a)\sim \mathcal{D}, \{y_i\}_{i=1}^K\sim \pi_{\theta_\text{old}}(\cdot\mid q)}\\&
\Bigg[ 
    \frac{1}{K} \sum_{i=1}^{K} \frac{1}{|y_i|} \sum_{t=1}^{|y_i|} 
    \Bigg( 
        \min \Big( r_{i,t}(\theta) {A}_{i,t}, \ \ \text{clip} \Big( r_{i,t}(\theta), 1 - \varepsilon, 1 + \varepsilon \Big) {A}_{i,t} \Big)- \beta D_{\text{KL}}(\pi_{\theta} || \pi_{\text{ref}}) 
\Bigg) \Bigg],
\label{eq:grpo_loss}
\end{aligned}
\end{equation}
where $r_{i,t}(\theta)=\frac{\pi_{\theta}(y_{i,t} \mid q, y_{i,<t})}{\pi_{\theta_{\text{old}}}(y_{i,t} \mid q,y_{i,<t})}$ and $\varepsilon$ is the clip bound. $D_{\text{KL}}(\pi_{\theta} || \pi_{\text{ref}})$ is the token-level KL loss, keeping the policy model $\pi_\theta$ not far from the reference policy $\pi_{ref}$.

\subsection{C3PO: Constrained Contextual Computation Policy Optimization}
\label{sec:c3po}
\paragraph{Observation.}
While long-CoT reasoning remains essential for handling complex tasks, current GRPO methods demonstrate systemic length-related issues that fundamentally undermine training stability, especially for the long-CoT reasoning models. Specifically, our analysis reveals two following critical issues affecting the training stability:
\begin{itemize}
    \item {\textbf{Within-step length bias:}} Within a single batch, unpacked responses of varying lengths induce substantial gradient bias under the GRPO objective, as initially identified in pioneering studies~\citep{yu2025dapo,liu2025understanding}.
    This bias primarily arises from the length-normalized gradient estimation by normalizing per-response rewards through division by their token counts (termed per-token scaling), the procedure systematically amplifies gradient magnitudes for shorter sequences while attenuating them for longer ones. While recent research~\citep{yu2025dapo,liu2025understanding} has introduced token-level loss mechanisms to address within-step length bias, the persistence of across-step gradient variance continues to pose challenges, as elaborated below. 
    
    \item 
    \textbf{Across-step gradient variance:} During RL training with exploratory mechanisms, the policy model's generated responses exhibit substantial stochastic variance in sequence length.
    This dynamic sequence length variation induces non-trivial optimization challenges for token-level optimizers, as fluctuating lengths create training token inconsistencies that propagate through the learning pipeline. 
    As empirically validated in Figure~\ref{fig:MoeLite-Dynamics}, highly-variation length fluctuation (Figure~\ref{fig:moelite-length}) results in an abnormal gradient characteristics (Figure~\ref{fig:moelite-gradient-norm}), ultimately lead to premature reward collapse (Figure~\ref{fig:moelite-rewards}). 
\end{itemize}
In addition to the challenges associated with training instability, our empirical observations revealed that variations in response length significantly influence training efficiency. Specifically, longer response sequences result in increased inference and training latency, whereas shorter sequences compromise computational throughput efficiency, as shown in Figure~\ref{fig:moelite-throughput}.

\begin{figure*}[ht]
\centering
\includegraphics[width=0.8\textwidth]{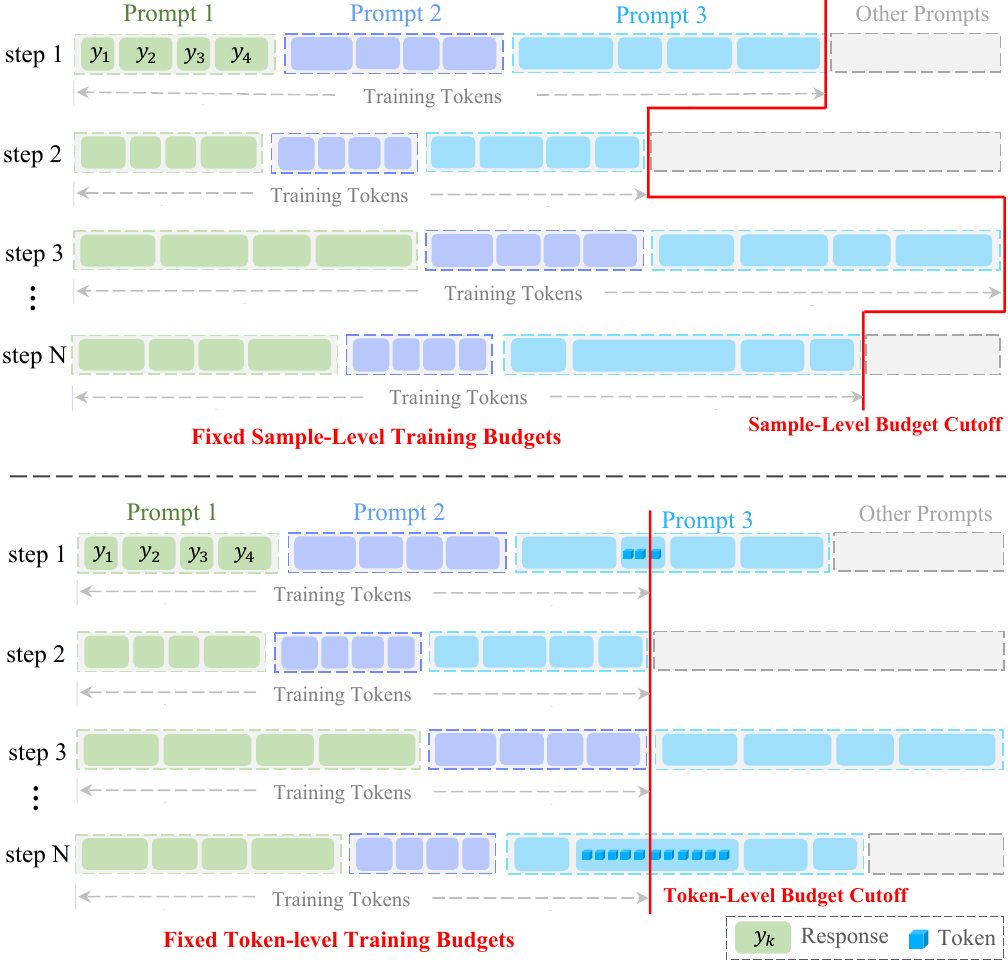}
\caption{\textbf{The comparison between our C3PO strategy and the widely-used dynamic sampling strategy, The C3PO performs token truncation when the token count exceeds the budget, after advantage computation but prior to gradient backpropagation}}
\label{fig:c3po-overview}
\end{figure*}

\paragraph{Methodology.}
Building upon these empirical observations, we posit that synergistic algorithm-engineering co-design constitutes a foundational requirement for achieving stable and scalable reinforcement learning training. To translate this principle into practice, we introduce \underline{C}onstrained \underline{C}ontextual \underline{C}omputation \underline{P}olicy \underline{O}ptimization(\textbf{C3PO}), an innovative token-level optimization framework designed to mitigate training instability while enhancing throughput consistency. 
The core innovation lies in establishing a formalized computational budgeting system that imposes explicit constraints on gradient contributions at the token level, thus ensuring homogeneous gradient contributions across variable-length sequences. 
With the training token budget, the GRPO loss function in Eq. \ref{eq:grpo_loss} is reformulated as follows,
\begin{equation}
\begin{aligned}
\mathcal{L}_\text{C3PO}(\theta)& = -\mathbb{E}_{\{q,a\}_{l=1}^{L}\sim \mathcal{D},{\{y_i\}_{i=1}^K\sim \pi_{\theta_\text{old}}(\cdot\mid q)}} \\&
\Bigg[ 
    \frac{1}{\Phi} \sum_{i=1}^{|\mathcal{S}|} \sum_{t=1}^{|y_i|} 
    \textcolor{red}{\mathbb{I}\Bigg[y_{i,t} \in \Psi\Bigg]}
    \Bigg( 
        \min \Big( r_{i,t}(\theta) {A}_{i,t}, \ \ \text{clip} \Big( r_{i,t}(\theta), 1 - \varepsilon, 1 + \varepsilon \Big) {A}_{i,t} \Big)- \beta D_{\text{KL}}(\pi_{\theta} || \pi_{\text{ref}}) 
\Bigg) \Bigg], \\&
    \textcolor{red}{\text{s.t.}\quad |\Psi| = \Phi}
\label{eq:c3po_loss}
\end{aligned}
\end{equation}
where $\Phi$ is the training token budget (i.e., constrained contextual computation), $\Psi$ is the selected tokens by custom sampling strategy. $\mathcal{S}$ is the selected responses for training, $L$ is the query size per step and $K$ denotes the group size, $|y_i|$ denotes the token size for $i-$th responses. In our experiment the set $\Psi$ is sampled as below:
\begin{equation}
\begin{aligned}
\Psi = \{ (y_1, y_2, \cdots, y_N) | y_i \in \mathcal{B}\}, \quad \text{s.t.} \; N \leq |\mathcal{B}|, \sum_{j=1}^{N-1} |y_j| < \Phi, \;
 \sum_{j=1}^{N} |y_j| \geq \Phi
\end{aligned}
\end{equation}

where $\mathcal{B}=\{(q,a);\{y_i\}_{i=1}^{K}\}_{l=1}^{L}$ is the entire set in a training step.
During a practical training step, we implement token-budgeted dynamic sampling in the training phase. 
For each batch, we employ a greedy algorithm to iteratively select sufficient responses $\mathcal{S}$ whose cumulative token count closely exceeds or equals the predefined token budget $\Phi$. 

By implementing a fixed token budget per optimization step, C3PO systematically mitigates GRPO's sensitivity to variations in individual sequence lengths. This design facilitates two critical improvements: 1) Homogeneous Gradient Scaling: The uniform factor $1/\Phi$ ensures equivalent gradient contributions across responses of varying token lengths, resolving the disproportionate weighting bias between short and long sequences inherent in conventional approaches. Furthermore, such a design mitigates abnormal gradient magnitudes caused by fluctuations in response length, effectively preventing the destabilization of training dynamics and subsequent reward collapse; 2) Deterministic Training Dynamics: Predictable computational loads eliminate burst-induced latency spikes while ensuring step-time consistency in distributed training environments.

Complementing the C3PO framework, we utilize entropy regularization~\citep{he2025skywork} to the loss function, explicitly penalizing low-variance action distributions and thereby encouraging exploration of the policy model.
\begin{equation}
\mathbb{H}(\theta) = \mathcal{H}(\pi_\theta(\cdot\mid y_{i,<t})).
\end{equation}

Figure~\ref{fig:c3po-overview} is the comparsion of our C3PO Strategy with the widely-used Dynamic Sampling Strategy~\citep{yu2025dapo}. Each line represents a batch of grouped responses generated from the policy model for a single training step. In each group, there are the same number of responses, each composed of variant tokens. In each training step, all tokens are aggregated to form a token level global batch, which is then fed into optimizers such as AdamW~\citep{loshchilov2019decoupledweightdecayregularization}. Due to the variable length of total tokens, the optimizer faces challenges as the gradients exhibit high variance, resulting in convergence difficulties. Previous approaches, such as dynamic sampling~\citep{yu2025dapo,tencenthunyuanteam2025hunyuanturbosadvancinglargelanguage}, operate at the sample level by filtering and removing samples, yet fail to adequately address this class of problems. While C3PO operates at the token level by sampling tokens to form a token level global batch, each training step maintains consistent token input to optimizer. This results in reduced gradient variance and consequently achieves significantly more stable optimization.

Our model \model{} adopts MoE architecture, which fundamentally differs from conventional dense models due to its inclusion of multiple specialized experts. However, this design introduces challenges related to expert imbalance during training. To enhance training efficiency and to prevent imbalances in token distributions across experts, we incorporate load balance loss and router z-loss detailed in ~\cite{ling}, which is formulated as:
\begin{equation}
    \mathbb{B}(\theta) = \frac{1}{N_e} \sum_{i=1}^{N_e} P_i * F_i * N_e,\quad
    \mathbb{Z}(\theta) = \frac{1}{M}\sum_{i=1}^{M} \Big(\log \sum_{j=1}^{N_e} \exp (z_{ij})\Big)^2,
\end{equation}
where $N_e$ is the number of experts, $M$ is the number of tokens, $P_i$ and $F_i$ are the average probability and count of the $i$-th expert selected across all tokens in a batch, and $z_{ij}$ is the logits of the router.

In summary, the overall loss for training is formulated as:
\begin{equation}
\begin{aligned}
    \mathcal{L} = \mathcal{L}_\text{C3PO}(\theta) + \alpha_{entropy} * \mathbb{H}(\theta) + \alpha_{balance} * \mathbb{B}(\theta) + \alpha_{zloss} * \mathbb{Z}(\theta)
\end{aligned}
\end{equation}

\subsection{Reward Model}
\subsubsection{Math \& Science Verifier}
Our training framework incorporates rule-based verifiable rewards in reinforcement learning, which has been proven effective for advancing the reasoning ability of large language models~\citep{deepseekai2025deepseekr1}. For mathematical and scientific tasks, we append a brief instruction prompt after each input query to facilitate long chain-of-thought reasoning, \textit{i.e.}, \texttt{Please reason step by step, and put your final answer within \textbackslash\textbackslash boxed\{\}.} We employ external verification tool, \textit{i.e.}, Math-Verify~\footnote{https://github.com/huggingface/Math-Verify}, to evaluate the correctness of model responses. Specifically, a score of $1$ is awarded for responses that correctly match the ground-truth answers, and a score of $0$ is assigned to incorrect solutions. Since Math-Verify provides robust parsing ability that well accommodates various mathematical notations and expressions, we do not include any explicit format-related reward in our training framework.

\subsubsection{Code Verifier}
For code task, we build a code sandbox for reward verification, supporting code execuation and online judgeg tasks across several programming languages (e.g., Python, C++, Java). It offers multiple execution modes (function calls, online judging, unit testing) and interaction paradigms (real-time SDK/API for training, offline batch processing for data cleaning), achieving 8K/s throughput with sub-second latency. With the code sandbox, we employs a sparse outcome reward for RL training on code tasks. Specifically, the reward is defined based on the execuation results from the sandbox, \textit{i.e.},
\[
R_{\text{code}} = \left\{
\begin{array}{ll}
1,\qquad \text{All test cases passed}\\
0,\qquad \text{Otherwise}
\end{array}
\right.
\]
It's worth noting that the reward mechanism employs a sparse design, wherein a reward of 1 is exclusively granted only if the code successfully passes all test cases; otherwise, the reward remains 0. This approach stands in stark contrast to incremental reward systems that offer partial credit for solutions that are incomplete or only partially correct. By adopting this strategy, we ensure that models are incentivized to gain a thorough understanding of the problem, rather than focusing on superficial test cases. This prevents models from simply regurgitating answers to public test cases or overfitting to trivial edge cases, encouraging a more robust and well-rounded approach to problem-solving.

\subsection{Training Pipeline}
\begin{figure*}[ht]
\centering
\includegraphics[width=0.7\textwidth]{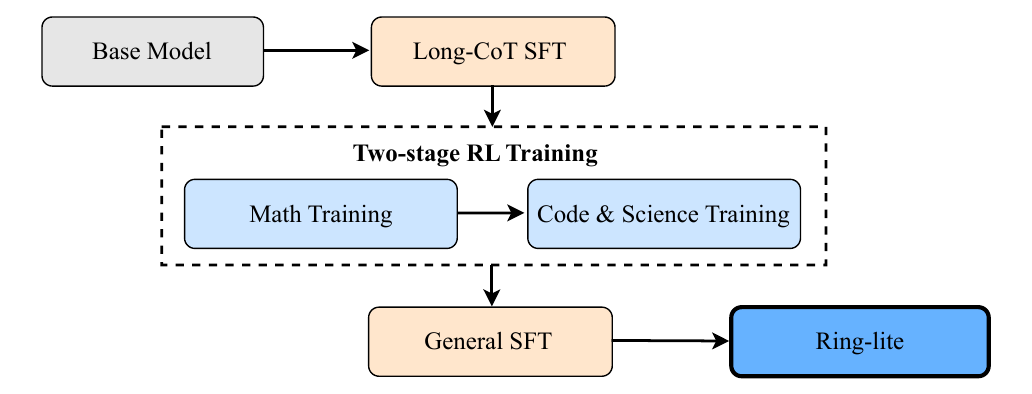}
\caption{\textbf{The training pipeline of Ring-lite}}
\label{fig:ring-rl-pipeline}
\end{figure*}

The overview of the training pipeline is depicted in Figure~\ref{fig:ring-rl-pipeline}. It consists of a four-stage training process. We first conduct Long Chain-of-Thought Supervised Fine-Tuning (Long-CoT SFT) to obtain our distilled model, \textit{i.e.}, \distillmodel{}. This stage aims to directly distill the reasoning ability of a larger teacher model into our small-sized base model. From our preliminary analysis, we find that with our meticulously curated reasoning data, the reasoning ability of the distilled model can be further enhanced. However, directly applying reinforcement learning (RL) on mixed reasoning data is vulnerable to domain conflict, resulting in performance declines. Thus, we propose adopting a two-stage RL training pipeline: first, we run RL training on a math dataset, then incorporate code and science datasets in subsequent RL training. This approach empirically demonstrates that it can effectively preserve reasoning abilities across diverse fields. As both Long-Cot SFT and the two-stage RL training focus on improving performance on reasoning tasks, we additionally include a general SFT stage to enhance the model's ability in various general tasks, such as instruction following, creative writing, safety, and etc.

%% file: sections/exp.tex
\section{Experiment}
\label{sec:bench}
\subsection{Experimental Setup}
\subsubsection{Training Settings}
As introduced in the training pipeline, to enhance the model's reasoning capabilities, we performed SFT on Ling-lite-1.5\footnote{https://huggingface.co/inclusionAI/Ling-lite-1.5}~\citep{ling} using the well-constructed Long-CoT dataset. For optimization, we employed the AdamW optimizer with a weight decay of 0.1 and a learning rate of 3e-4, following a cosine decay schedule with a 1\% linear warmup. The training configuration included a batch size of 256 over 3 epochs. To facilitate long-context reasoning, we set the context window of the model to 32,768 tokens and adjusted the RoPE base to 600,000 for improved stability.

In RL training with C3PO, we use a batch size $L$ of 512, sampling $K=16$ responses per prompt and adopt a learning rate of $3e-6$ with AdamW optimizer. The token-budget parameter is set to $409600$. The maximum total length is configured to $24576$ and is extended to $32768$ in the second stage of code \& science training.
We set entropy loss coefficient $\alpha_{ent}=5e-4$, load balance loss coefficient $\alpha_{balance}=1e-5$, router z-loss coefficient $\alpha_{zloss}=1e-7$, KL loss coefficient $\beta=1e-3$.
All experiments were performed on 256 * NVIDIA H800.
\input{sections/benchmark}
\input{sections/baselines}
\input{sections/generation_setup}

\subsection{Main Results}
The evaluation results are shown in Table~\ref{tab:main_res}. 
To provide a fair comparison, we evaluate our \model{} against recent competitive reasoning models with approximately 10B parameters. As shown in Table~\ref{tab:main_res}, our \model{} achieves the best average score across multiple reasoning tasks while using only 2.75B active parameters. This establishes our \model{} as the new state-of-the-art reasoning model among small-scale MoE models, offering performance comparable to or even surpassing that of the latest strong reasoning dense model under 10B parameters, \textit{i.e.}, Qwen3-8b-Thinking. Additionally, compared to our previously released distilled MoE model, Ring-lite-distill-preview, our \model{} significantly improves reasoning performance on all benchmarks, further demonstrating the superiority of our training pipeline.
\begin{table*}[ht!]
\setlength\tabcolsep{2pt}
\def\w{20pt} 
\caption{%
\centering
  The comparison of different reasoning models. The best and second-best results are in \textbf{bold} and \underline{underlined}. Results with $^*$ are collected from their original papers. All models are evaluated with  the same evaluation setting.}
\label{tab:main_res}
\vspace{-8pt}
\centering\footnotesize
\scalebox{1}{
\begin{tabular}{l@{\extracolsep{10pt}}l@{\extracolsep{10pt}}c@{\extracolsep{10pt}}c@{\extracolsep{10pt}}c@{\extracolsep{10pt}}c|@{\extracolsep{10pt}}c>{}c@{\extracolsep{10pt}}c}
\toprule
 & &  \makecell{\textbf{R1-Distill} \\ \textbf{-Qwen-7B}}  &  \makecell{\textbf{R1-Distill} \\ \textbf{-Qwen-14B}} & \makecell{\textbf{AceReason} \\ \textbf{-Nemotron-7B}}  &  \makecell{\textbf{Qwen3-8b} \\ \textbf{-Thinking}} &  \makecell{\textbf{Ring-lite} \\ \textbf{-distill-preview}} & \textbf{\model{}} \\
\midrule
& Architecture & Dense & Dense & Dense & Dense & MoE & MoE \\
& \# Activated Params & 7B & 14B & 7B & 8B & 2.75B & 2.75B \\
& \# Total Params & 7B & 14B & 7B & 8B & 16.8B & 16.8B \\
\midrule
\multirow{5}*{\textit{Math}} & MATH500$_{\textcolor{gray}{pass@1}}$ & 92.80$^*$ & 93.90$^*$ & 94.10$^*$ &\textbf{97.40}$^*$ &  93.55 & \underline{96.65} &\\
 & AIME24$_{\textcolor{gray}{pass@1}}$ & 55.50$^*$ & 69.70$^*$ & 69.00$^*$ & \underline{76.00}$^*$ &  57.81 & \textbf{76.61} & \\
  & AIME25$_{\textcolor{gray}{pass@1}}$ & 39.43 & 52.29 & 53.60$^*$ & \underline{67.30}$^*$ & 39.38 & \textbf{69.11} & \\
 & CNMO24$_{\textcolor{gray}{pass@1}}$ & 62.50 & 70.14 & 71.61 & \underline{74.57} & 60.68 & \textbf{75.61} &\\
 & LiveMathBench$_{\textcolor{gray}{pass@1}}$ & 74.62 & 79.13 & 79.37 & \underline{81.90} & 72.37 & \textbf{83.74} & \\
\midrule
\multirow{2}*{\textit{Coding}} & LiveCodeBench$_{\textcolor{gray}{pass@1}}$ & 37.60$^*$ & 53.10$^*$ & 51.80$^*$ & \underline{58.06} &  32.62 & \textbf{60.66} & \\
  & CodeForces$_{\textcolor{gray}{percentile}}$ & 46.55 & 80.02 & \underline{84.03} & 79.44 & 48.62 & \textbf{86.45} & \\
\midrule
\multirow{2}*{\textit{Science}} & GPQA Diamond$_{\textcolor{gray}{pass@1}}$ & 49.10$^*$ & 59.10$^*$ & 50.67 & \textbf{62.00}$^*$ & 52.81 & \underline{61.05} &  \\
& OlympiadBench$_{\textcolor{gray}{pass@1}}$ & 70.26 & 74.86 & 75.95 & \underline{79.65} & 68.74 & \textbf{79.95} &  \\
\midrule
& Average & 58.71 & 70.25 & 70.01 & 75.15 & 58.51 & \textbf{76.65} &  \\
\bottomrule
\end{tabular}
}
\end{table*}

\input{sections/findings}

%% file: sections/benchmark.tex
\subsubsection{Benchmarks}
For a comprehensive evaluation of the quality of our reasoning models, we implemented automatic benchmarks to assess their performance, which are categorized into the following dimensions.
\begin{itemize}
\item \textbf{Math: } MATH-500~\citep{math-500}, AIME 2024, AIME 2025~\citep{aime}, CNMO 2024\footnote{https://www.cms.org.cn/Home/comp/comp/cid/12.html}, Livemathbench~\citep{livemathbench}, MinervaMath~\citep{minervamath}.
\item \textbf{Coding: } LiveCodeBench~\citep{livecodebench}(202408 – 202501), Codeforces\footnote{The Codeforces was assessed through problems from 14 Div. 2 contests of Codeforces, combined with expert-designed test cases, followed by the computation of expected ratings and competitor proportions.}.
\item \textbf{Science: } GPQA Diamond~\citep{gpqa}, OlympiadBench~\citep{OlympiadBench}.
\end{itemize}

%% file: sections/baselines.tex
\subsubsection{Baselines}
We conduct comprehensive evaluations against several baselines of similar parameter sizes, including
Qwen3-8B-Thinking~\citep{qwen3}, R1-Distill-Qwen-7B, R1-Distill-Qwen-14B~\citep{deepseekai2025deepseekr1}, AceReason-Nemotron-7B~\citep{AceReason-Nemotron} and Ring-lite-distill-preview~\citep{ringlitedistill}. 

%% file: sections/generation_setup.tex
\subsubsection{Evaluation Settings}
For all reasoning models, we utilize a temperature of 1.0, a top-p value of 0.95, and a maximum output length of 32,768 tokens. In addition, the prompts are unified with the zero-shot setting. For mathematics benchmarks, we use Math-Verify as the evaluator to score model generations.

%% file: sections/findings.tex
\subsection{Key Findings}
\label{sec:findings}
In this section, we analyze the central observations derived from training reinforcement learning across diverse domains, with particular focus on emergent instability phenomena, training efficiency between SFT and RL methodologies, and inter-domain data conflicts. 
To ensure a fair comparison, we applied supervised fine-tuning to the Qwen2.5-7B-Base~\citep{qwen25techreport} model using our Long-CoT dataset as stated in Section~\ref{sec:long-cot-data}, resulting in the \qwenbasedistill{} model.

\subsubsection{Move Towards Stable and Efficient RL}
As outlined in the methodology, we empirically observe significant training instability and throughput fluctuations in GRPO during RL training. Here we present a systematic experimental evaluation confirming these phenomena, followed by a quantitative analysis establishing the effectiveness of our proposed method in addressing both challenges:

\begin{itemize}[leftmargin=10pt, label={}]

\item \textbf{1. Reward Collapse Phenomenon in RL Training on Distilled Models}

During reinforcement learning training with distilled models, we found that reward trajectories exhibit a precipitous decline after a few training steps, failing to recover to baseline levels and culminating in complete training collapse. Through rigorous empirical diagnostics, we identify two critical factors governing RL training stability: \textbf{Model Entropy}, which quantifies policy degradation in distilled models, and \textbf{Response Length Fluctuation}, a measure of sequence generation instability. These factors demonstrate strong correlation with reward collapse, as evidenced by quantitative ablation studies as follows:

\begin{figure*}[!hbt]
\centering
\begin{subfigure}[t]{0.47\textwidth}
\centering
\includegraphics[width=\linewidth]{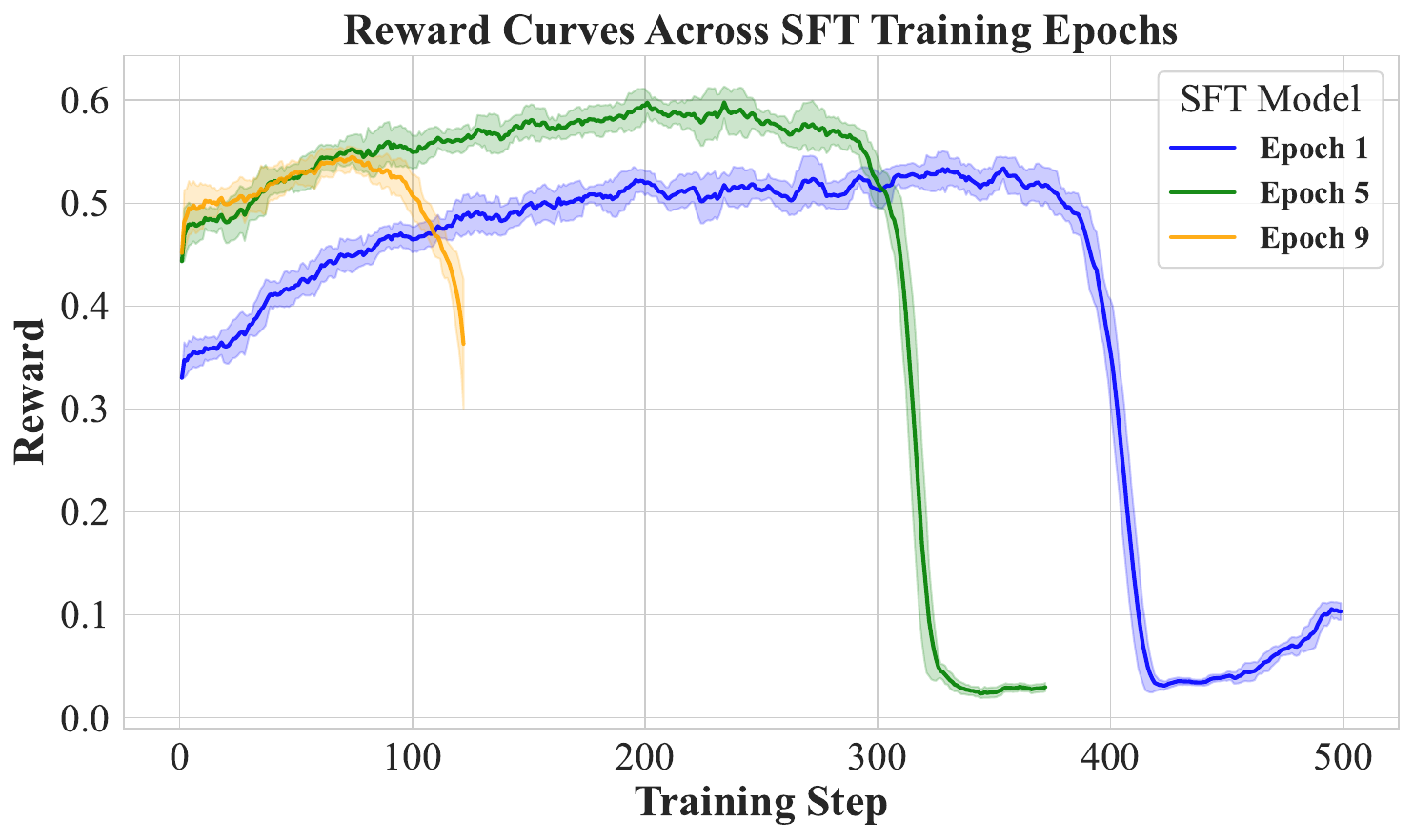}
\caption{\small\textbf{Ring-lite Reward}}
\label{fig:moe-diff-eps}
\end{subfigure}
\hfill
\centering
\begin{subfigure}[t]{0.47\textwidth}
\centering
\includegraphics[width=\linewidth]{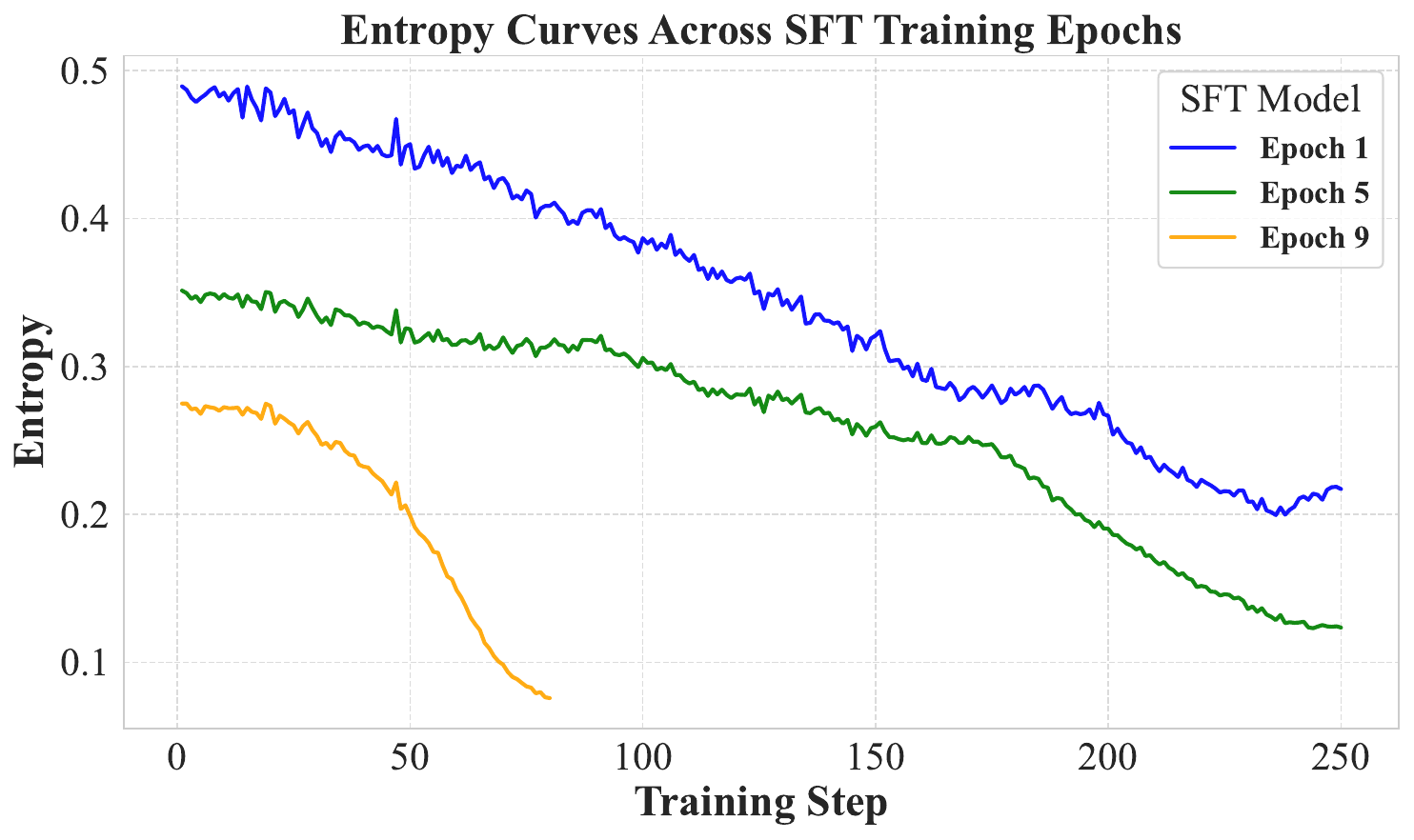}
\caption{\small\textbf{Ring-lite Entropy Loss}}
\label{fig:moelite-entropy-loss}
\end{subfigure}
\caption{\textbf{The reward and entropy curves of Ring-lite.}}
\label{fig:diff-eps}
\end{figure*}

\begin{figure*}[!hbt]
\centering
\begin{subfigure}[t]{0.47\textwidth}
\includegraphics[width=\linewidth]{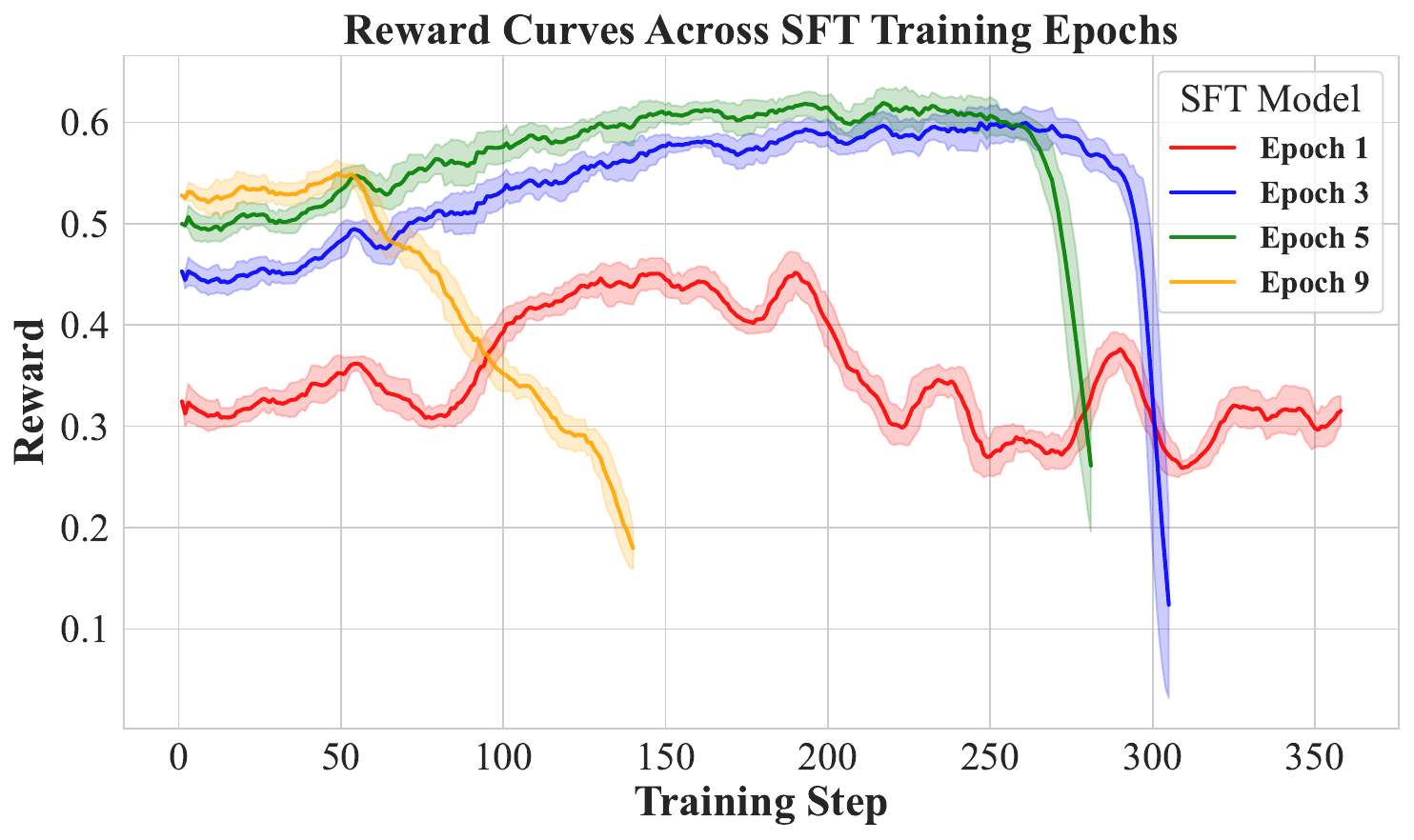}
\caption{\small\textbf{Ring-distill-Qwen-7B Reward}}
\label{fig:qwen-diff-eps}
\end{subfigure}
\hfill
\begin{subfigure}[t]{0.47\textwidth}
\centering
\includegraphics[width=\linewidth]{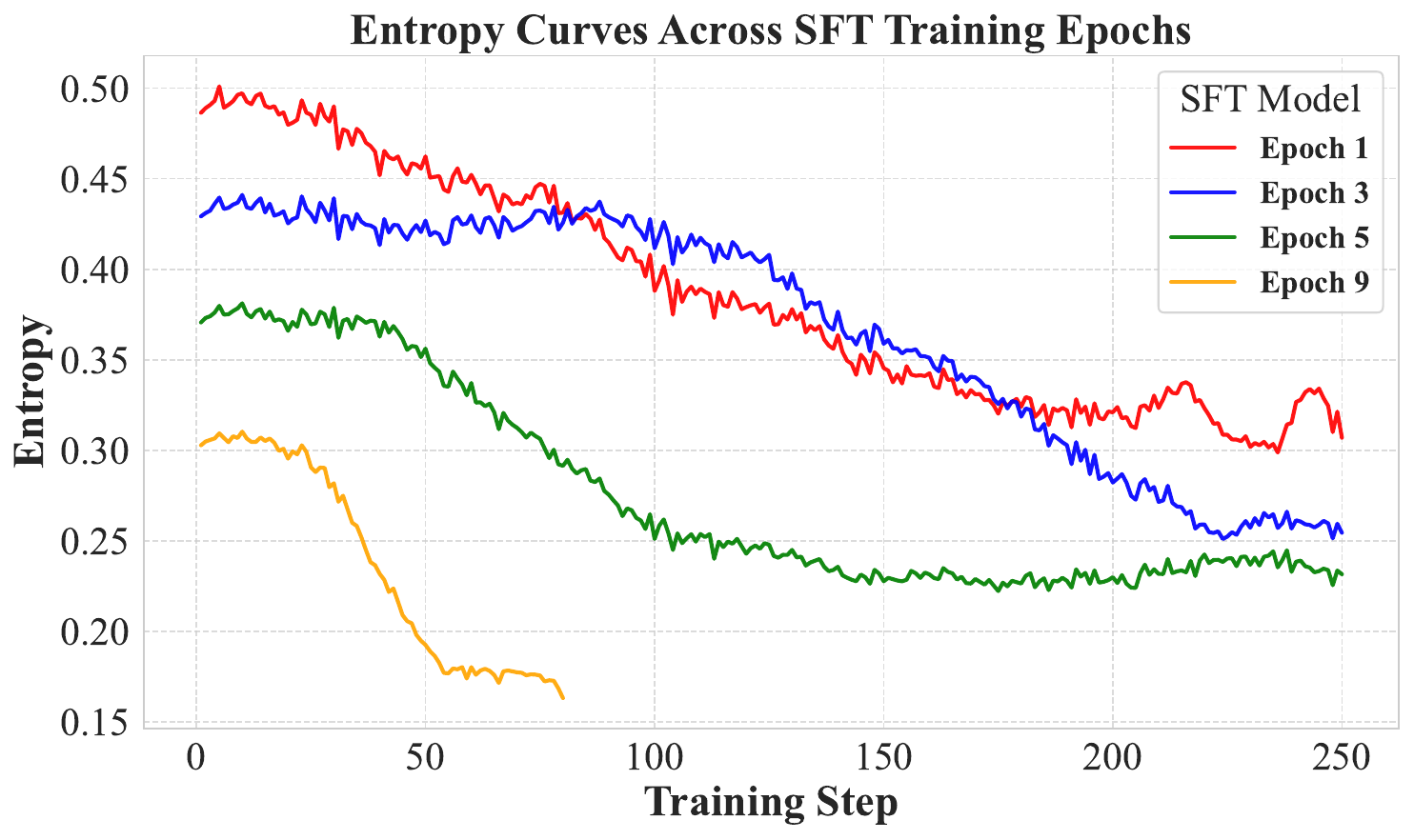}
\caption{\small\textbf{Ring-distill-Qwen-7B Entropy Loss}}
\label{fig:qwen-entropy-loss}
\end{subfigure}
\caption{\textbf{The reward and entropy curves of Ring-distill-Qwen-7B}}
\label{fig:entropy-loss}
\end{figure*}

\begin{itemize}
    \item \textbf{Model Entropy:} As shown in Figure~\ref{fig:diff-eps}, the reward collapse during RL training exhibits a systematic dependence on the number of Long-CoT SFT epochs: models trained with more SFT epochs experience collapse earlier. Specifically, models trained with a greater number of SFT epochs exhibit earlier onset of reward collapse. This trend is accompanied by a concurrent reduction in entropy loss (Figure~\ref{fig:entropy-loss}), revealing a robust inverse correlation between the magnitude of entropy loss and stability during RL training. Collectively, these results underscore that lower entropy loss during SFT corresponds to a higher propensity for reward collapse in subsequent RL phases, suggesting a statistically significant inverse relationship between these variables. Notably, this pattern persists across both MoE and dense model architectures, indicating architectural invariance in the observed phenomena.
    \item \textbf{Response Length Fluctuation:} Figure~\ref{fig:MoeLite-Dynamics} (Ring-lite) and Figure~\ref{fig:Qwen-Dynamics} (Ring-distill-Qwen-7B) demonstrate that the generation length exhibits great variability across training steps, resulting in significant fluctuations in training token sizes. These unstable token training sizes greatly affect optimization stability, as evidenced by pronounced increases and occasional spikes—in gradient norms, leading to catastrophic reward collapse. This observation underscores the imperative need for developing strategies that mitigate both entropy loss and generation length variability to ensure stable RL training.
\end{itemize}

\begin{figure*}[!hbt]
\centering
\begin{subfigure}[t]{0.48\textwidth}
\centering
\includegraphics[width=\linewidth]{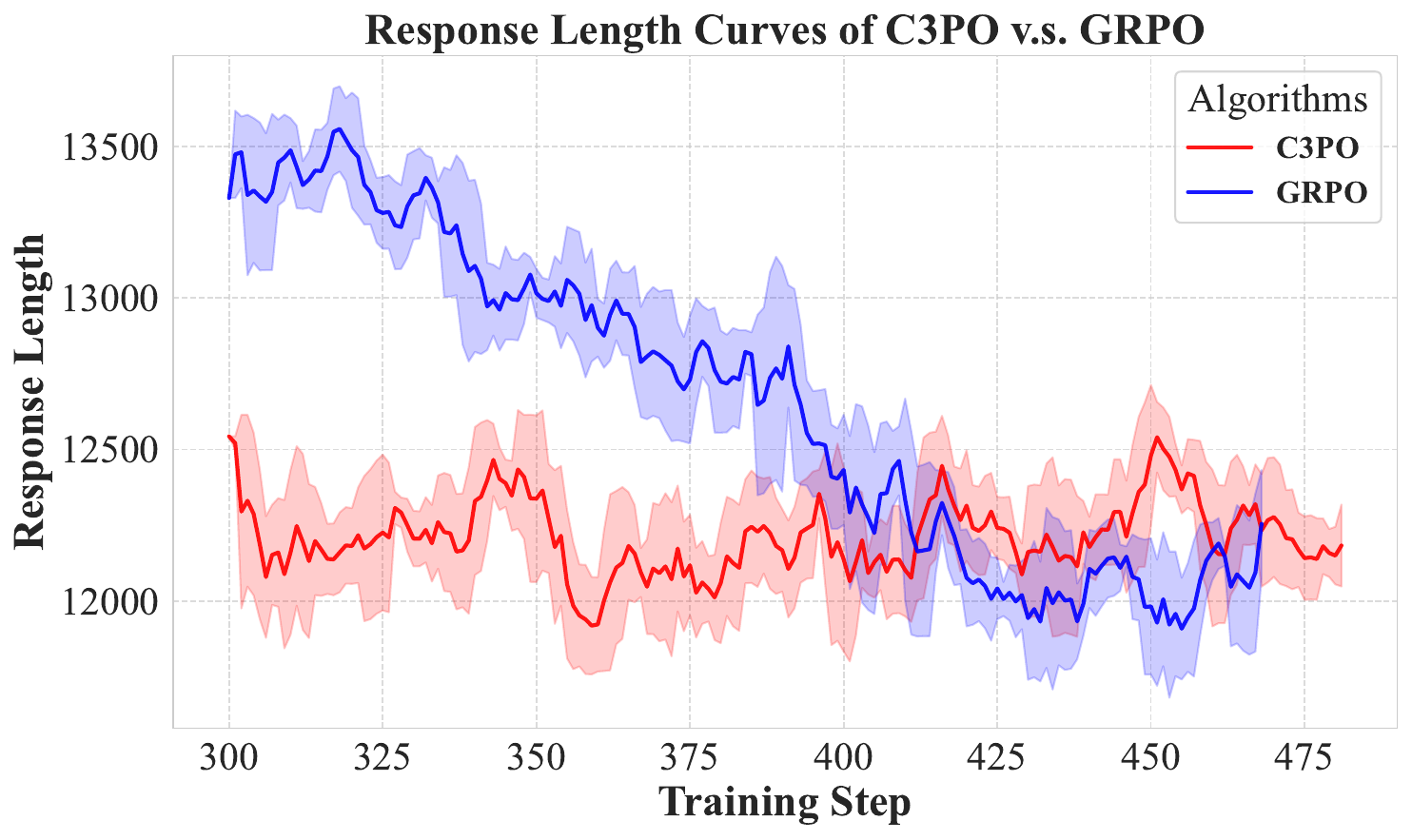}
\caption{\small\textbf{Response-Length}}
\label{fig:moelite-length}
\end{subfigure}
\hfill
\begin{subfigure}[t]{0.48\textwidth}
\centering
\includegraphics[width=\linewidth]{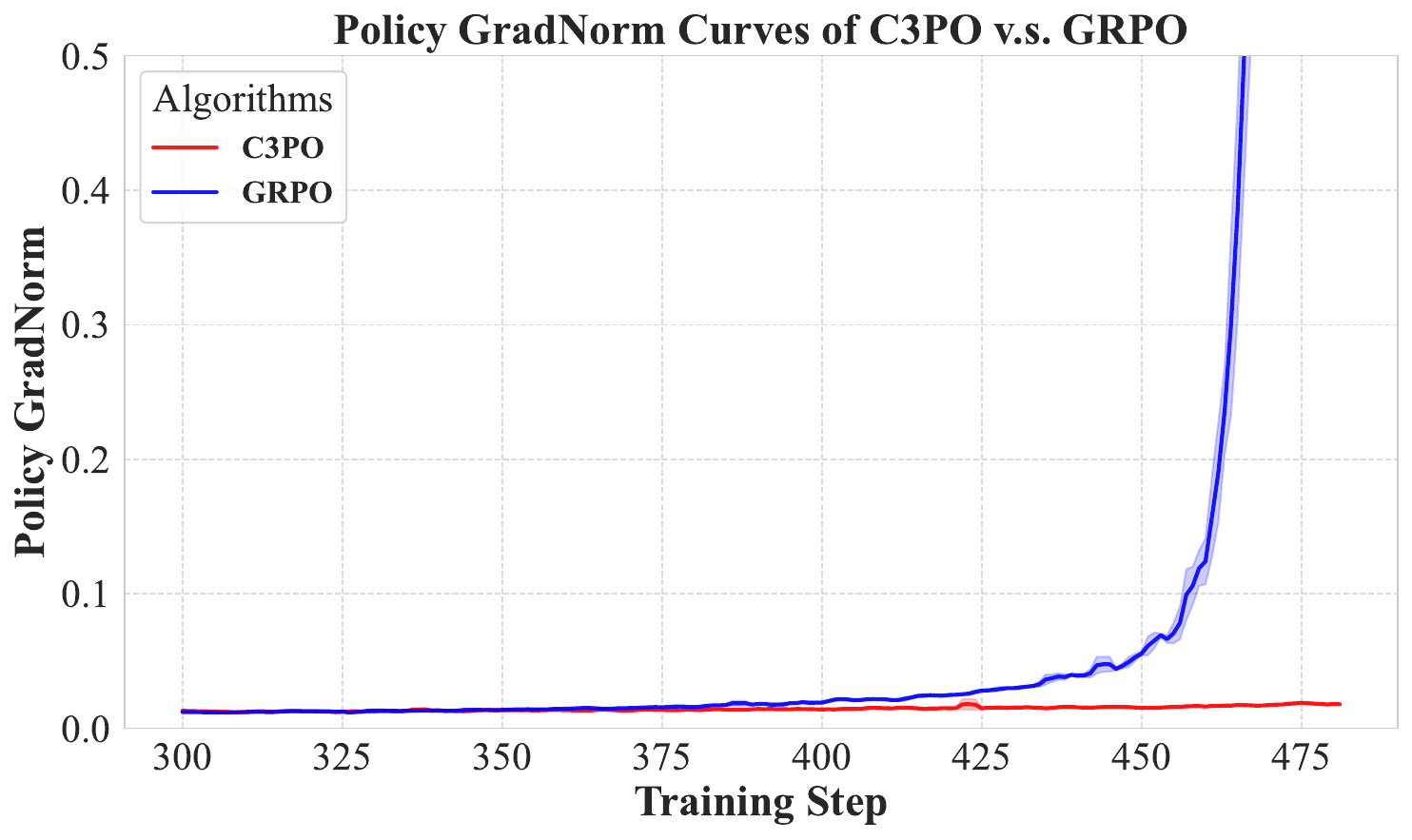}
\caption{\small\textbf{Gradient-Norm}}
\label{fig:moelite-gradient-norm}
\end{subfigure}
\begin{subfigure}[t]{0.48\textwidth}
\centering
\includegraphics[width=\linewidth]{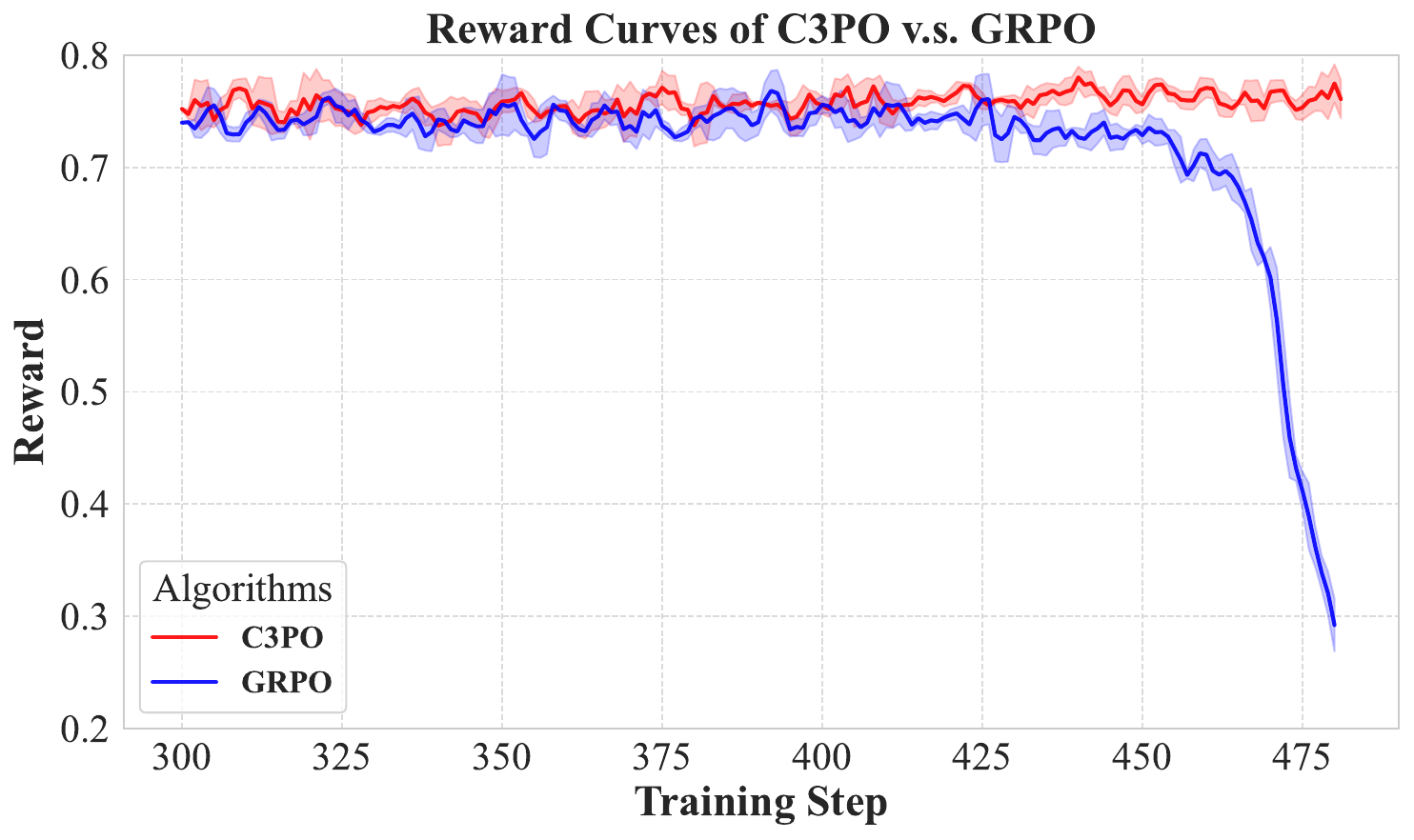}
\caption{\small\textbf{Rewards}}
\label{fig:moelite-rewards}
\end{subfigure}
\hfill
\begin{subfigure}[t]{0.48\textwidth}
\centering
\includegraphics[width=\linewidth]{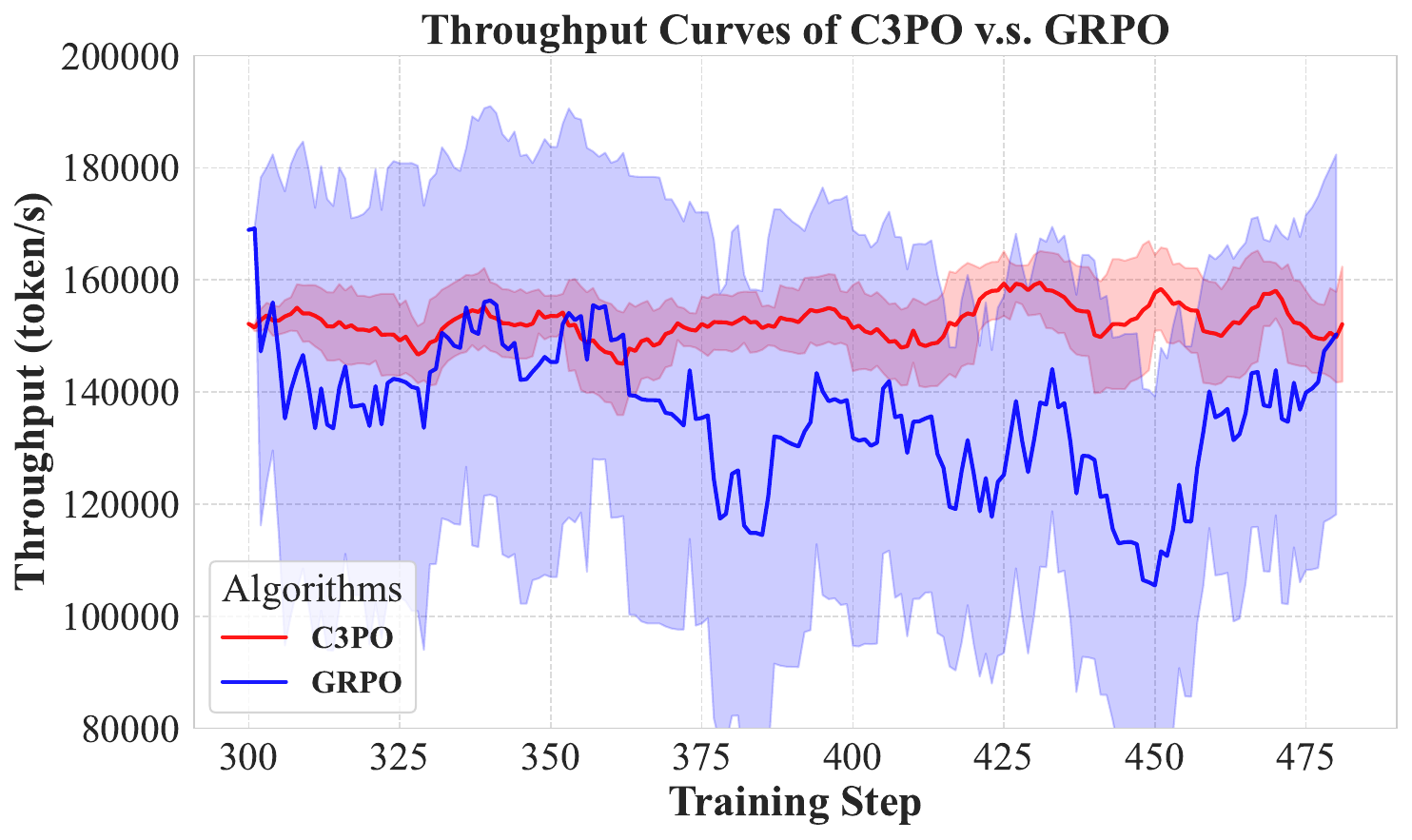}
\caption{\small\textbf{Throughput}}
\label{fig:moelite-throughput} 
\end{subfigure}
\caption{\textbf{Training Dynamics of Ring-lite}}
\label{fig:MoeLite-Dynamics}
\end{figure*}

\begin{figure*}[!hbt]
\centering
\begin{subfigure}[t]{0.48\textwidth}
\centering
\includegraphics[width=\linewidth]{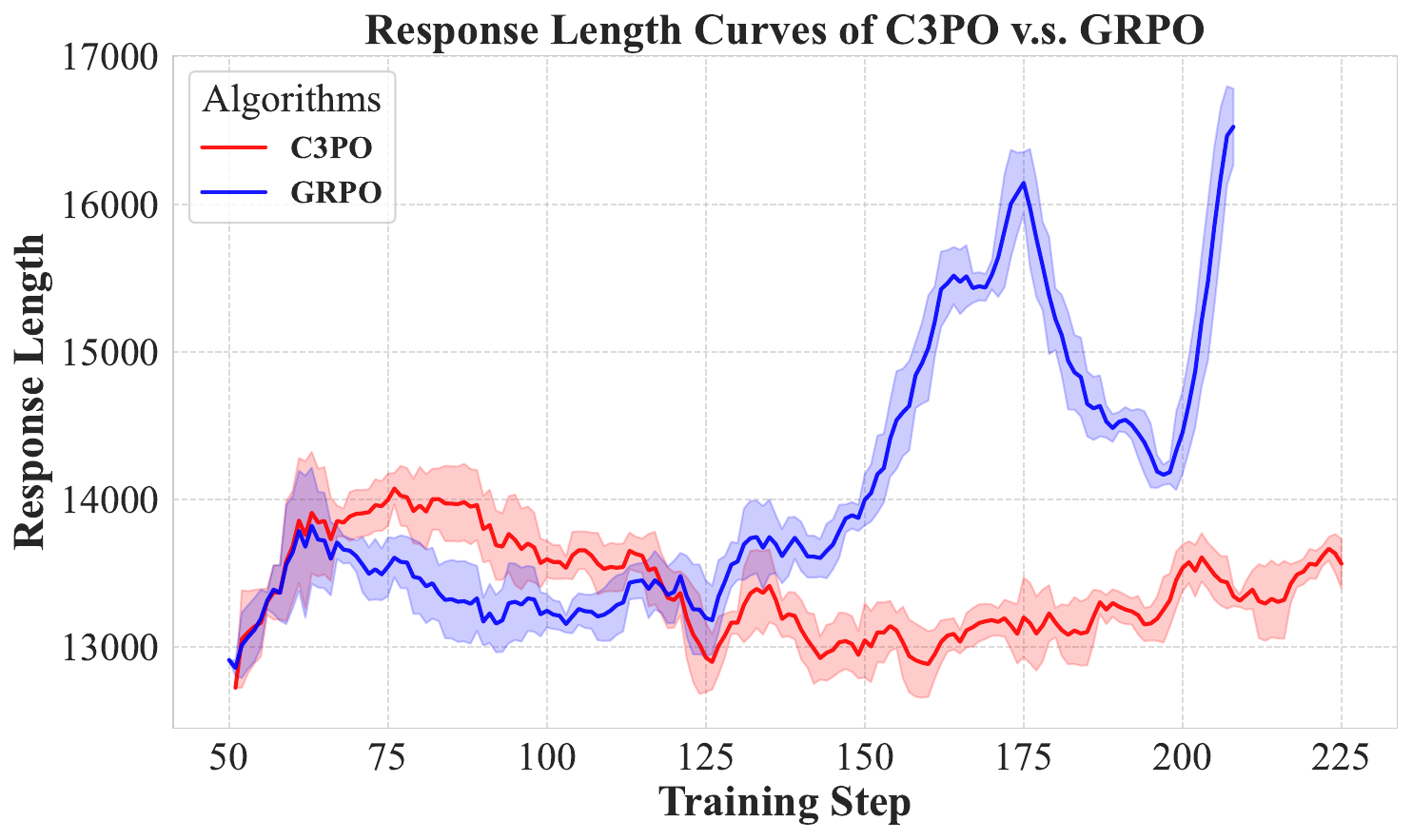}
\caption{\small\textbf{Response-Length}}
\label{fig:qwen-length}
\end{subfigure}
\hfill
\begin{subfigure}[t]{0.48\textwidth}
\centering
\includegraphics[width=\linewidth]{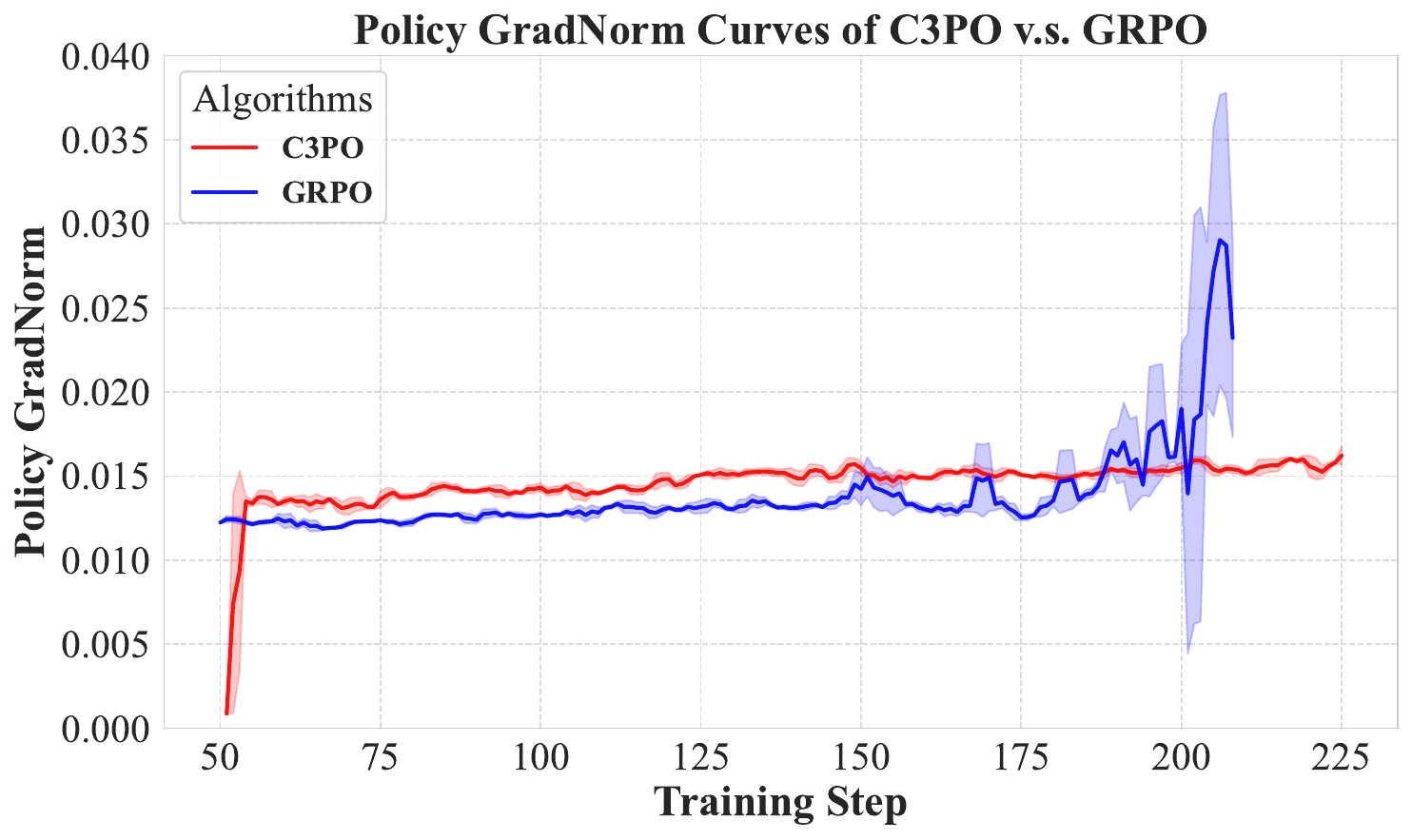}
\caption{\small\textbf{Gradient-Norm}}
\label{fig:qwen-gradient-norm}
\end{subfigure}
\begin{subfigure}[t]{0.48\textwidth}
\centering
\includegraphics[width=\linewidth]{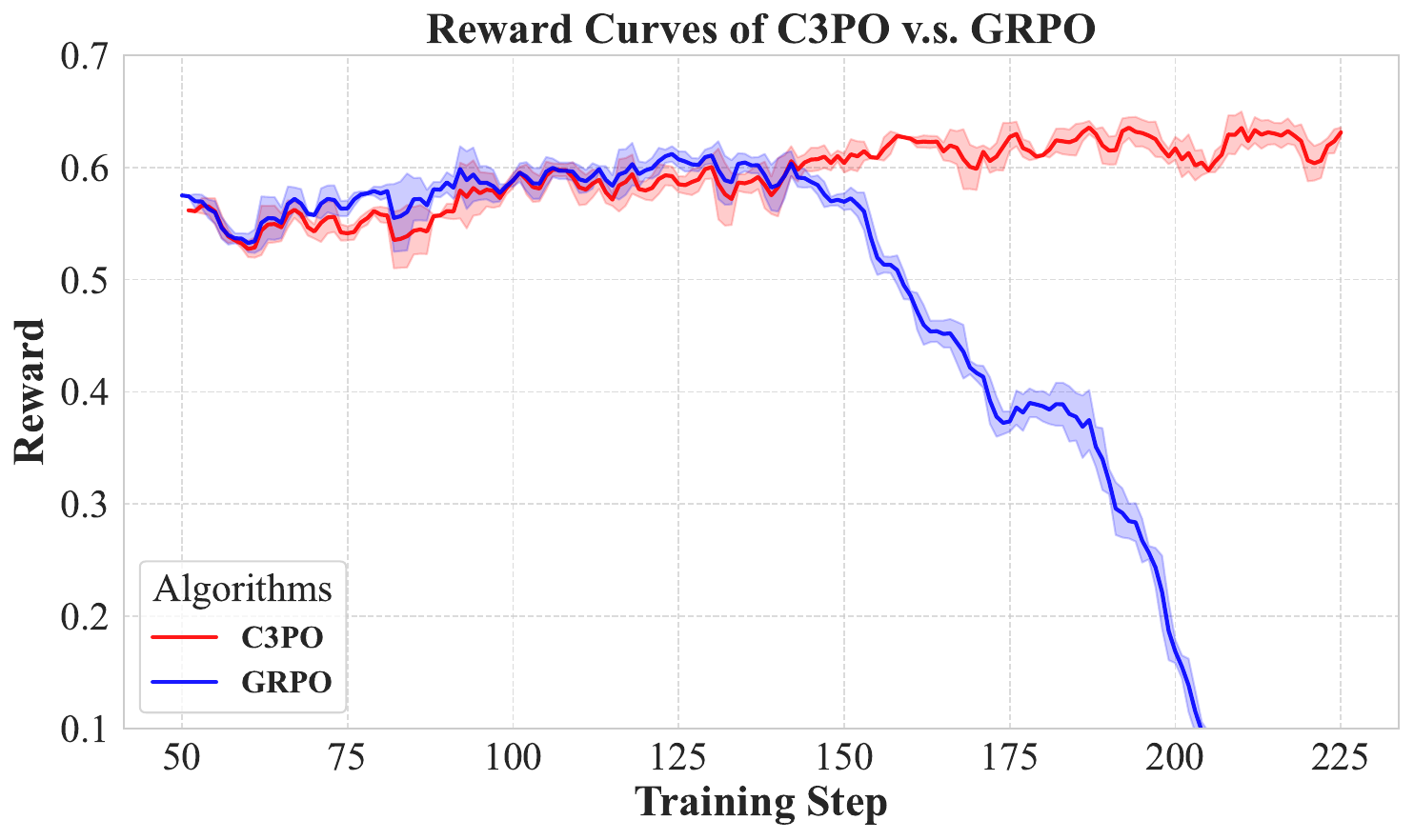}
\caption{\small\textbf{Rewards}}
\label{fig:qwen-rewards}
\end{subfigure}
\hfill
\begin{subfigure}[t]{0.48\textwidth}
\centering
\includegraphics[width=\linewidth]{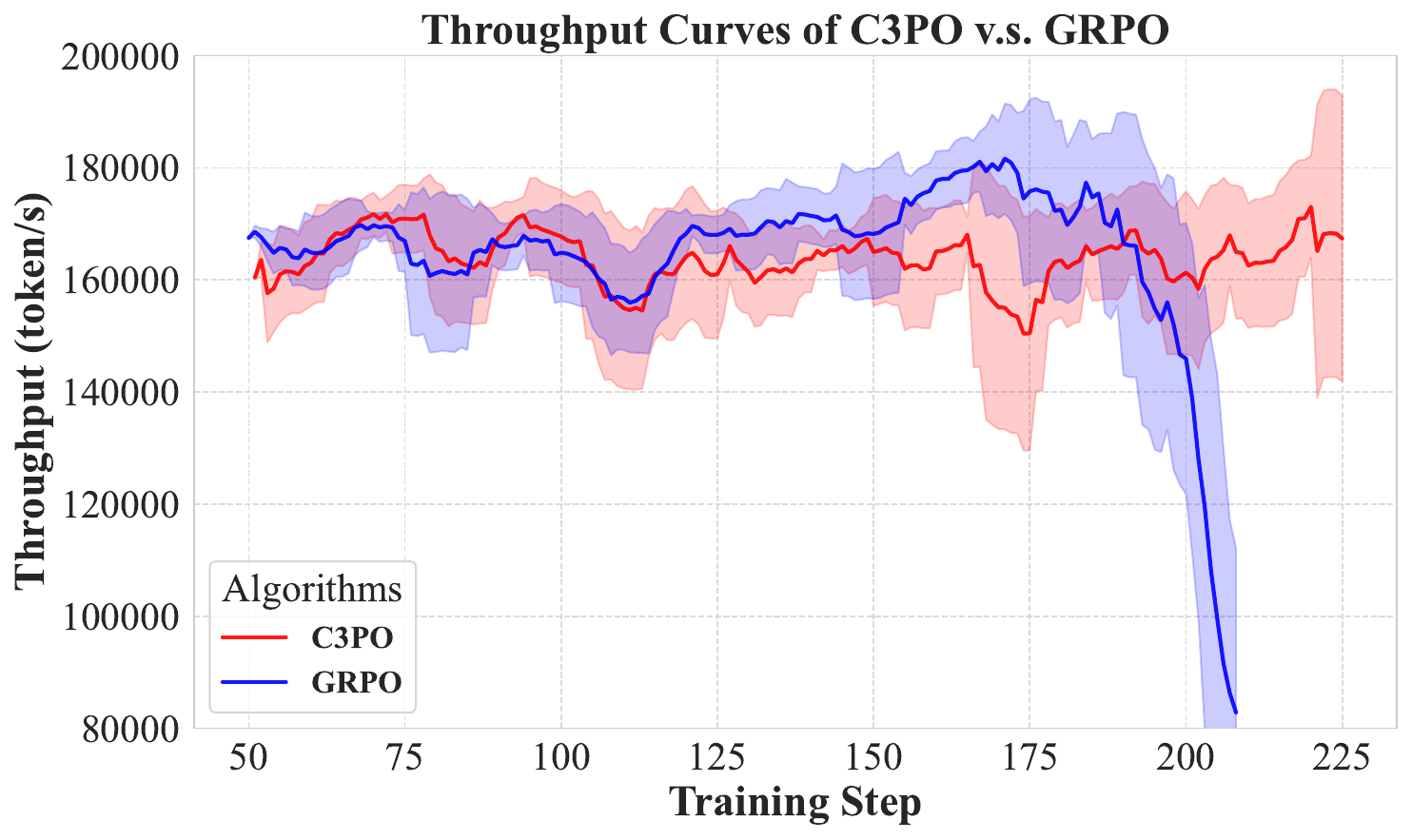}
\caption{\small\textbf{Throughput}}
\label{fig:qwen-throughput} 
\end{subfigure}
\caption{\textbf{Training Dynamics of Ring-distill-Qwen-7B}}
\label{fig:Qwen-Dynamics}
\end{figure*}

\item \textbf{2. RL Training Throughput Fluctuation}

Besides the challenge of reward collapse, our empirical observations reveal substantial throughput fluctuations emerging during RL training, which present considerable challenges for optimizing training efficiency in distributed systems. As shown in Figure \ref{fig:MoeLite-Dynamics} (Ring-lite) and Figure \ref{fig:Qwen-Dynamics} (Ring-distill-Qwen-7B), these variations are primarily attributable to response length variability. Specifically, longer response sequences necessitate prolonged computation per training step, whereas shorter sequences underutilize computational resources, thereby diminishing throughput efficiency. This dynamic throughput behavior introduces significant optimization challenges in system design, as unpredictable computational demands complicate the implementation of efficient resource allocation and scheduling.

\end{itemize}

Since GRPO method suffers from the training instability and throughput fluctuation problems, C3PO is proposed to address these limitations through two key mechanisms: (1) selecting SFT checkpoints associated with higher entropy loss to stabilize optimization dynamics, and (2) imposing a fixed token budget during training to ensure training token consistency. To validate the efficacy of our method, we conduct experiments using two distilled initialization models with different model architectures: Ring-lite-distill and Ring-distill-Qwen-7B.

As illustrated in Figures~\ref{fig:MoeLite-Dynamics} and~\ref{fig:Qwen-Dynamics}, models trained with fewer epochs exhibit enhanced stability during RL training. However, our empirical analysis (Table \ref{tab:sft_epochs}) reveals a critical trade-off between training stability and final model performance. While models initialized with Epoch 1 checkpoints demonstrate superior stability in both configurations, their performance metrics lag significantly behind those of Epoch 7 and Epoch 3. Conversely, Epoch 9 achieves the highest initial performance but suffer from destabilization during later RL training phases, ultimately failing to surpass the results of Epoch 7 and Epoch 3. Furthermore, our methodological innovation in maintaining fixed training token size enables C3PO to consistently outperform GRPO across four critical metrics: generation length stability, gradient stability, reward stability, and throughput stability (Figures \ref{fig:MoeLite-Dynamics} and \ref{fig:Qwen-Dynamics}). In short, our C3PO not only resolves the instability inherent in GRPO on distilled models but also ensures efficient RL training, thereby bridging the gap between training robustness and model capability.

\subsubsection{From Distill to RL: The Art of Balancing Token Efficiency}
For our experiments, we used a constant warm-up learning rate scheduler with rates [1e{-6}, 3e{-6}], using the AdamW optimizer. Specifically, Ring-MoE performed better with a learning rate of 3e{-6}, while Qwen achieved better results at 2e{-6}. We utilized a training batch size of 512 prompts, a minibatch size of 2, and generated 16 responses for each prompt. In both the rollout and evaluation phases, the temperature was set to 1.0 to promote response diversity.
For all methods, we set the bound $\epsilon$ to 0.2, the KL coefficient was set to 0.001. The maximum total length is configured to 24576. All experiments were performed on 256 * NVIDIA H800.

To further validate the generalizability of our findings, we conducted additional experiments on the Qwen series of models~\citep{qwen}, which have emerged as highly influential in the open-source community. Following the same methodology applied to \model{}, we first fine-tuned the Qwen2.5-7B-Instruct model using our Long-CoT dataset and subsequently performed RL training. We find that while distillation is effective, it requires significantly more training tokens than reinforcement learning (RL) to achieve comparable performance. Empirically, selecting checkpoints based on entropy loss within the range of 0.3{-0.5} yields optimal results on our RL training setting. Entropy loss values below this threshold limit model exploration and reduce the chances of learning to solve more challenging problems, whereas excessive entropy loss leads to slower convergence and degraded model performance.

\begin{figure*}[ht]
\centering
\begin{subfigure}[t]{0.49\textwidth}
\centering
\includegraphics[width=\linewidth]{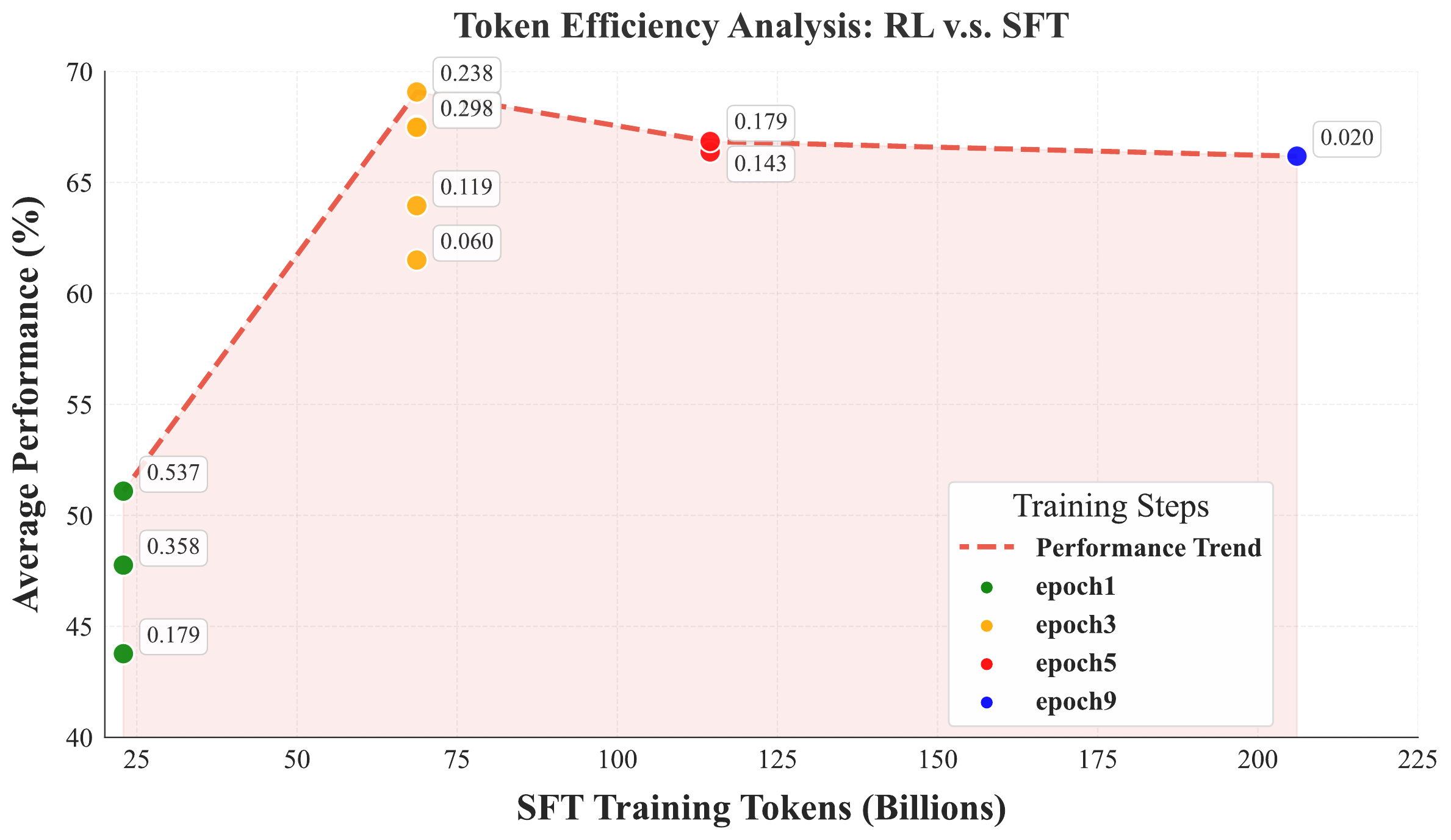}
\caption{\small\textbf{The token efficiency on Ring-distill-Qwen-7B.}}
\label{fig:qwen-token-efficiency}
\end{subfigure}
\hfill
\begin{subfigure}[t]{0.49\textwidth}
\centering
\includegraphics[width=\linewidth]{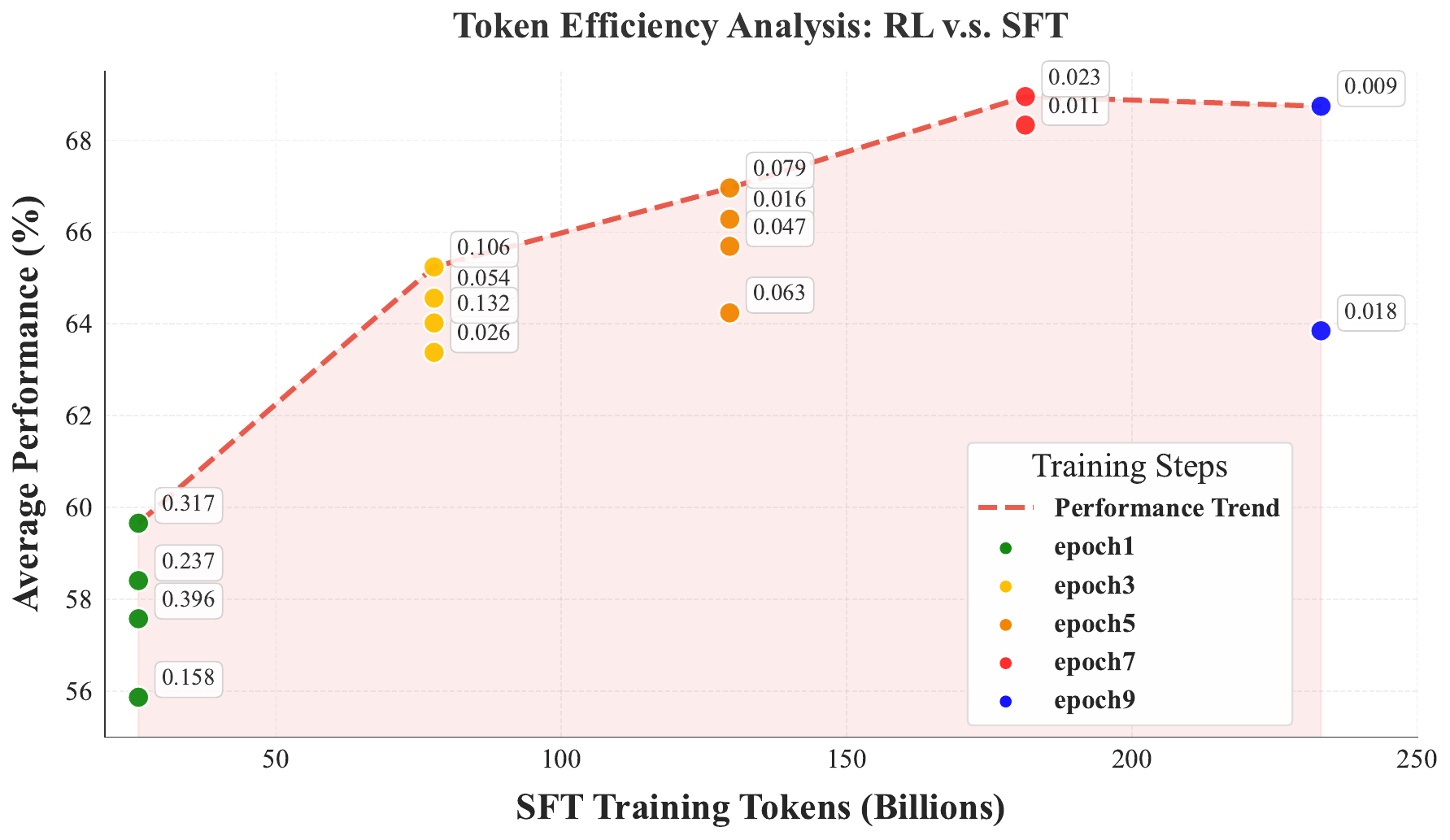}
\caption{\small\textbf{The token efficiency analysis on Ring-lite.}}
\label{fig:token-efficiency}
\end{subfigure}
\caption{\textbf{Reward curves across different SFT training epochs: (a) Ring-distill-Qwen-7B, (b) Ring-lite.} Dots with the same color denote different values of token efficiency on the same SFT model.}
\label{fig:combined-diff-eps}
\end{figure*}

From Figure~\ref{fig:moelite-entropy-loss}, we observe that varying the number of training epochs of the distilled model significantly influences the trend of entropy loss, thereby determining the exploration scope for RL. Based on our experiments, increasing the number of SFT training epochs leads to a rapid collapse of entropy. However, insufficient SFT training inevitably results in inferior performance. To systematically quantify the choice of optimal SFT epoch, we employ token efficiency, \textit{i.e.},  $\frac{\text{\# RL training tokens}}{\text{\# SFT training tokens}}$, to evaluate the relationships among RL training steps, SFT training steps, and average downstream performance. As shown in Figure~\ref{fig:combined-diff-eps} and Table~\ref{tab:sft_epochs}, the best performance is achieved with a moderate number of SFT training epochs and suitable token efficiency. In our training pipeline, we utilize these findings to select the optimal SFT model.

\begin{table*}[hbt!]
\setlength\tabcolsep{2pt}
\def\w{20pt} 
\caption{%
\centering
  The comparison of RL training across different SFT epochs. The best results are in \textbf{bold}. $\bigtriangleup_{impr}$ denotes the average performance improvements compared to respective SFT models.
}
\vspace{-8pt}
\centering\footnotesize
\scalebox{1}{
\begin{tabular}{l@{\extracolsep{10pt}}l@{\extracolsep{10pt}}l@{\extracolsep{10pt}}l@{\extracolsep{10pt}}l@{\extracolsep{10pt}}l|@{\extracolsep{10pt}}l@{\extracolsep{10pt}}c}
\toprule
\textbf{Model} & \textbf{SFT Epochs} & \textbf{AIME24@32} & \textbf{AIME25@32} & \textbf{GPQA} & \textbf{LiveCodeBench} & \textbf{Average} & $\bigtriangleup_{impr}$\\
\midrule
\multirow{3}*{\qwenbasedistill{}} & Epoch 1 & 32.97 & 31.67 & 45.33 & 36.42 & 36.60 & +10.19  \\
& Epoch 3 & 62.45 & 49.01 & 57.17 & 48.61 & \textbf{54.31} & \textbf{+11.03} \\
& Epoch 5 & 61.51 & 50.42 & 57.29 & 45.25 & 53.62 & +5.62  \\
& Epoch 9 & 56.35 & 44.64 & 56.22 & 46.86 & 51.02 & +0.13 \\
\midrule
\multirow{3}*{\model{}} & Epoch 1 & 51.35 & 38.91 & 53.6 & 40.32 & 46.05 & +3.03 \\
& Epoch 5 & 63.54 & 42.08 & 53.88 & 44.13 & 50.91 & \textbf{+8.76}  \\
 & Epoch 7 & 64.74 & 43.12 & 57.61 & 46.82 & \textbf{53.07} & +7.27  \\
& Epoch 9 & 65.73 & 43.85 & 56.91 & 45.70 & 53.05 & +7.93  \\
\bottomrule
\end{tabular}
}
\label{tab:sft_epochs}
\end{table*}

\subsubsection{Resolving Domain Data Conflict: Beyond Mixed Solutions}
For our experiments, we used a constant warm-up learning rate scheduler with rates [2e{-6}, 3e{-6}], using the AdamW optimizer. Specifically, Ring-MoE performed better with a learning rate of 3e{-6}, while Qwen achieved better results at 2e{-6}. We utilized a training batch size of 512 prompts, and generated 16 responses for each prompt. In both the rollout and evaluation phases, the temperature was set to 1.0 to promote response diversity.
For all methods, the KL coefficient was set to 0.001. The maximum total length is configured to 24576. All experiments were performed on 256 * NVIDIA H800.

In our preliminary reinforcement learning (RL) experiments, we observed significant performance declines across various reasoning benchmarks when training our cold-start supervised fine-tuning (SFT) model with a combination of math and code reasoning datasets. We then conducted RL experiments on two representative distilled dense models: \dsbasedistill{} and \qwenbasedistill{}.

As shown in Table~\ref{tab:dense_data_conflict}, combining reasoning datasets from the math and code domains does not lead to performance gains across different fields. Instead, the mixed dataset fails to outperform models trained exclusively on either math or code datasets. 
Notably, the experimental findings derived from our distilled models reveal that math-only training achieves superior performance on coding benchmarks compared to code-only, irrespective of the model's architectural configuration. However, this observation does not extend to the DeepSeek-derived models, indicating that the performance of RL training may be strongly influenced by the Long-CoT data in the SFT period.
Conversely, code-only RL does not provide additional improvements for math tasks. These results indicate that mixing diverse reasoning domains may introduce conflicts that hinder overall performance. Specialized training on individual domains appears to be more effective for optimizing performance in each specific area.

\begin{table*}[ht!]
\setlength\tabcolsep{2pt}
\def\w{20pt} 
\caption{%
\centering
  The comparison of different training stages on Ring-lite, Qwen and DeepSeek distilled model. The best results are in \textbf{bold}, the performance differences compared to the best performance are denoted with arrows and numbers.
}
\vspace{-8pt}
\centering\footnotesize
\scalebox{1.1}{
\begin{tabular}{l@{\extracolsep{10pt}}l@{\extracolsep{10pt}}l@{\extracolsep{10pt}}l@{\extracolsep{10pt}}l@{\extracolsep{10pt}}l}
\toprule
\textbf{Model} & \textbf{Training Stages} & \textbf{AIME24@32} & \textbf{AIME25@32} & \textbf{GPQA} & \textbf{LiveCodeBench} \\
\midrule
\multirow{3}*{\dsbasedistill{}} & Math-only & \textbf{59.64} & \textbf{44.22} & \textbf{49.94} & 39.74\\
& Code-only & 55.62 & 41.77 & 48.93 & \textbf{42.56} \\
& Math \& Code Mixed & 58.44\textcolor{blue}{$\downarrow_{1.20}$} & 43.80\textcolor{blue}{$\downarrow_{0.42}$} & 49.81\textcolor{blue}{$\downarrow_{0.13}$} & 42.20\textcolor{blue}{$\downarrow_{0.36}$} \\
\cmidrule{2-6}
\multirow{3}*{\qwenbasedistill{}} & Math-only & \textbf{66.77} & \textbf{57.60} & 58.18 & \textbf{50.72}\\
& Code-only & 55.21 & 45.68 & 55.30 & 47.94 \\
& Math \& Code Mixed & 62.86\textcolor{blue}{$\downarrow_{3.91}$} & 54.22 \textcolor{blue}{$\downarrow_{3.38}$} & \textbf{58.74} & 49.73\textcolor{blue}{$\downarrow_{0.99}$} \\
\cmidrule{2-6}
\multirow{3}*{\model{}} & Math-only & \textbf{78.54} & \textbf{67.60} & 61.61 & \textbf{61.11} \\
& Code-only & 73.12 & 64.74 & 61.27 & 59.27 \\
& Math \& Code Mixed & 76.77\textcolor{blue}{$\downarrow_{1.77}$} & 66.61\textcolor{blue}{$\downarrow_{0.99}$} & \textbf{61.80} & 60.44\textcolor{blue}{$\downarrow_{0.67}$} \\
\bottomrule
\end{tabular}
}
\label{tab:dense_data_conflict}
\end{table*}

To enhance overall reasoning performance across diverse areas when training with multiple domain-specific datasets, we divided the our RL training into multiple stages. Specifically, we first conducted RL experiments using only the math dataset, followed by applying RL with scientific and code datasets. As shown in Table~\ref{tab:moe_data_conflict}, our two-stage training strategy significantly improved downstream performance on challenging reasoning benchmarks, such as AIME25 and LiveCodeBench. Additionally, by doubling the amount of code and scientific training data, we achieved an average performance increase of 1\% on both math and scientific benchmarks. Based on these results, we adopted this two-stage training strategy for \model{} to maintain superior overall reasoning abilities across multiple domains.

\begin{table*}[h]
\setlength\tabcolsep{2pt}
\def\w{20pt} 
\caption{%
\centering
  The comparison of different training strategies.
}
\vspace{-8pt}
\centering\footnotesize
\scalebox{1.2}{
\begin{tabular}{l@{\extracolsep{10pt}}l@{\extracolsep{10pt}}l@{\extracolsep{10pt}}l@{\extracolsep{10pt}}l}
\toprule
\textbf{Training Strategy} & \textbf{AIME24@32} & \textbf{AIME25@32} & \textbf{GPQA} & \textbf{LiveCodeBench} \\
\midrule
Naive Mixed RL & 76.20 & 64.06 & 61.71 & 60.04 \\
Two-stage RL & 77.19 \textcolor{red}{$\uparrow_{0.99}$} & 65.73\textcolor{red}{$\uparrow_{1.67}$} & 61.48\textcolor{blue}{$\downarrow_{0.23}$} & 61.87\textcolor{red}{$\uparrow_{1.83}$} \\
\midrule
\textit{Increase code \& stem data} \\
Naive Mixed RL & 75.94 & 65.31 & 60.54 & 59.81 \\
Two-stage RL & 78.54\textcolor{red}{$\uparrow_{2.60}$} & 67.71\textcolor{red}{$\uparrow_{2.40}$} & 61.93\textcolor{red}{$\uparrow_{1.39}$} & 61.78\textcolor{red}{$\uparrow_{1.97}$} \\
\bottomrule
\end{tabular}
}
\label{tab:moe_data_conflict}
\end{table*}
\FloatBarrier

%% file: sections/conclu.tex
\section{Conclusion}
\label{sec:conclu}
This work introduces \model{}, a novel reasoning MoE model that achieves state-of-the-art performance across diverse challenging tasks. By integrating the proposed reinforcement learning method C3PO, \model{} significantly enhances reasoning capabilities while maintaining computational efficiency and stability. Empirical results demonstrate its efficacy, achieving scores of 76.61\% (AIME24), 69.11\% (AIME25), 60.66\% (LiveCodeBench), 86.45\% (Codeforces) and 61.05\% (GPQA-diamond). Our research pursues two primary objectives: first, to refine RL training through algorithm-engineered co-design for enhanced efficiency; second, to address complex tasks requiring advanced reasoning to extend the boundaries of model intelligence.

In the future, we plan to investigate general reward modeling to attain human-verifier-level accuracy and explore systematic data synthesis techniques for scalable dataset expansion, thereby improving generalization capabilities.

\clearpage

%% file: sections/author.tex
\section{Contributors}
\label{sec:contri}

\DTLnewdb{names}
\DTLnewrow{names} \DTLnewdbentry{names}{name}{Zihao Wang}
\DTLnewrow{names} \DTLnewdbentry{names}{name}{Kuan Xu}
\DTLnewrow{names} \DTLnewdbentry{names}{name}{Jia Guo}
\DTLnewrow{names} \DTLnewdbentry{names}{name}{Xin Zhao}
\DTLnewrow{names} \DTLnewdbentry{names}{name}{Xiaoyun Feng}
\DTLnewrow{names} \DTLnewdbentry{names}{name}{Yingting Wu}
\DTLnewrow{names} \DTLnewdbentry{names}{name}{Liang Jiang}
\DTLnewrow{names} \DTLnewdbentry{names}{name}{Zhenyu Huang}
\DTLnewrow{names} \DTLnewdbentry{names}{name}{Shuaicheng Li}
\DTLnewrow{names} \DTLnewdbentry{names}{name}{Yongkang Liu}
\DTLnewrow{names} \DTLnewdbentry{names}{name}{Zhenduo Zhang}
\DTLnewrow{names} \DTLnewdbentry{names}{name}{Junwu Xiong}
\DTLnewrow{names} \DTLnewdbentry{names}{name}{Xinyu Kong}
\DTLnewrow{names} \DTLnewdbentry{names}{name}{Feng Zhu}
\DTLnewrow{names} \DTLnewdbentry{names}{name}{Xinxing Yang}
\DTLnewrow{names} \DTLnewdbentry{names}{name}{Zujie Wen}
\DTLnewrow{names} \DTLnewdbentry{names}{name}{Zhiqiang Zhang}
\DTLnewrow{names} \DTLnewdbentry{names}{name}{Jun Zhou}
\DTLnewrow{names} \DTLnewdbentry{names}{name}{Qing Cui}
\DTLnewrow{names} \DTLnewdbentry{names}{name}{Jiaming Liu}
\DTLnewrow{names} \DTLnewdbentry{names}{name}{Dingnan Jin}
\DTLnewrow{names} \DTLnewdbentry{names}{name}{Shaomian Zheng}
\DTLnewrow{names} \DTLnewdbentry{names}{name}{Wang Ren}
\DTLnewrow{names} \DTLnewdbentry{names}{name}{Deng Zhao}
\DTLnewrow{names} \DTLnewdbentry{names}{name}{Bin Hu}
\DTLnewrow{names} \DTLnewdbentry{names}{name}{Xiaodong Yan}
\DTLnewrow{names} \DTLnewdbentry{names}{name}{Cai Chen}
\DTLnewrow{names} \DTLnewdbentry{names}{name}{Liangcheng Fu}
\DTLnewrow{names} \DTLnewdbentry{names}{name}{Junbo Zhao}
\DTLnewrow{names} \DTLnewdbentry{names}{name}{Xiaopei Wan}
\DTLnewrow{names} \DTLnewdbentry{names}{name}{Yang Li}
\DTLnewrow{names} \DTLnewdbentry{names}{name}{Kaihong Zhang}
\DTLnewrow{names} \DTLnewdbentry{names}{name}{Zhenglei Zhou}
\DTLnewrow{names} \DTLnewdbentry{names}{name}{Lei Liang}
\DTLnewrow{names} \DTLnewdbentry{names}{name}{Tongkai Yang}
\DTLnewrow{names} \DTLnewdbentry{names}{name}{Ding Liu}
\DTLnewrow{names} \DTLnewdbentry{names}{name}{Hao Dai}
\DTLnewrow{names} \DTLnewdbentry{names}{name}{Jun Mei}
\DTLnewrow{names} \DTLnewdbentry{names}{name}{Qiang Gao}
\DTLnewrow{names} \DTLnewdbentry{names}{name}{Xuemin Yang}
\DTLnewrow{names} \DTLnewdbentry{names}{name}{Jiewei Wu}
\DTLnewrow{names} \DTLnewdbentry{names}{name}{Hongzhi Luan}
\DTLnewrow{names} \DTLnewdbentry{names}{name}{Zhankai Xu}
\DTLnewrow{names} \DTLnewdbentry{names}{name}{Quan Wan}
\DTLnewrow{names} \DTLnewdbentry{names}{name}{Longfei Zheng}

\DTLsort{name}{names}

\large{Authors are listed \textbf{alphabetically by the first name}.} 

\large{
\begin{multicols}{3}
\raggedcolumns
Ling Team\\
\DTLforeach*{names}{\thename=name}{\thename\\}
\end{multicols}}

\clearpage